\documentclass[sigconf]{acmart}
\AtBeginDocument{%
  }

\setcopyright{acmlicensed}
\copyrightyear{2026}
\acmYear{2026}
\setcopyright{cc}
\setcctype{by}
\acmConference[KDD '26]{Proceedings of the 32nd ACM SIGKDD Conference on Knowledge Discovery and Data Mining V.2}{August 09--13, 2026}{Jeju Island, Republic of Korea}
\acmBooktitle{Proceedings of the 32nd ACM SIGKDD Conference on Knowledge Discovery and Data Mining V.2 (KDD '26), August 09--13, 2026, Jeju Island, Republic of Korea}
\acmDOI{10.1145/3770855.3819070}
\acmISBN{979-8-4007-2259-2/2026/08}

\usepackage{amsfonts}
\usepackage{amsmath}
\usepackage{amsthm}
\usepackage{booktabs}
\usepackage{caption}
\usepackage{footnote}
\usepackage{url}
\usepackage{algorithm}
\usepackage{algpseudocode}
\usepackage{multirow}
\usepackage{array}

\theoremstyle{remark}
\newtheorem*{remark}{Remark}

\begin{document}

\title{A Framework for Evaluating and Benchmarking Concept Drift Detection Methods}

\author{Vitor Cerqueira}
\affiliation{%
  \institution{University of Coimbra}
  \city{Coimbra}
  \country{Portugal}
}
\email{vitorc@dei.uc.pt}

\author{Heitor Murilo Gomes}
\affiliation{%
  \institution{Victoria University of Wellington}
  \city{Wellington}
  \country{New Zealand}}
\email{heitor.gomes@vuw.ac.nz}

\author{Marco Heyden}
\affiliation{%
 \institution{Commerzbank}
 \city{Frankfurt}
 \country{Germany}}
\email{heydenmarco48@gmail.com}

\author{Bernhard Pfahringer}
\affiliation{%
  \institution{University of Waikato}
  \city{Waikato}
  \country{New Zealand}}
\email{bernhard@cs.waikato.ac.nz}

\author{Albert Bifet}
\affiliation{%
  \institution{AI Institute, University of Waikato}
  \city{Waikato}
  \country{New Zealand}}
\email{albert.bifet@telecom-paris.fr}

\renewcommand{\shortauthors}{Cerqueira et al.}

\begin{abstract}
Data stream mining is fundamentally challenged by concept drift, where distributional changes can degrade model performance. 
Despite the proliferation of drift detection methods, progress in the field is hindered by inconsistent evaluation practices: studies rely on oversimplified synthetic data generators, adopt incompatible metrics, and lack transparency in hyperparameter selection, making fair comparisons difficult. 
We address this gap with a novel benchmarking framework comprising three contributions: (1) a drift simulation method that injects controlled distributional changes into real-world datasets via Monte Carlo trials, enabling supervised evaluation while preserving real-world data complexity; (2) an evaluation protocol for drift detection with timing-aware criteria, including the derivation of new metrics (e.g., F1 detection score, normalized detection time) that are comparable across streams; and (3) we advocate for a leave-one-dataset-out hyperparameter optimization protocol for drift detection methods that promotes configuration robustness across heterogeneous stream dynamics. 
We benchmark 14 widely used drift detection methods on 7 real-world datasets across 4 drift types (class prior, label swap, feature permutation, feature filtering), each under both abrupt and gradual transitions. Our experimental results provide insights into the strengths and weaknesses of current drift detection approaches while establishing baseline performance metrics for future research in this area. All code and experiments are publicly available.

\end{abstract}




\maketitle

\section{Introduction}\label{sec:introduction}

Machine learning models typically rely on the assumption that the data encountered during inference follows the same distribution as the training data. However, when this distribution shifts, a phenomenon known as concept drift, model performance can degrade significantly, leading to inaccurate predictions. Therefore, reliably detecting and adapting to concept drift is crucial for maintaining the effectiveness and reliability of machine learning systems.

Detecting concept drift has long been recognized as a critical challenge in machine learning, motivating the development of numerous drift detection algorithms over the years. Notable examples include sequential analysis methods such as CUSUM and the Page-Hinkley test~\cite{page1954continuous}, control chart-based approaches such as DDM~\cite{gama2004learning} and EDDM~\cite{baena2006early}, and adaptive windowing methods such as ADWIN~\cite{bifet2007learning}.
Yet, despite extensive research on drift detection, there is still a lack of unified evaluation practices and reproducible baselines in the field. We identify two key sources of inconsistency:
\begin{enumerate}
    \item[I1.] \textbf{Methodology:} Existing studies are inconsistent w.r.t.\ evaluation metrics and protocols, hyperparameter selection strategies, datasets, and drift types.  
    \item[I2.] \textbf{Availability of labeled data:} Drift points---the times at which concept drifts occurred---are typically unknown in real-world data. Hence, evaluating a detector on real-world data must happen \emph{unsupervised}. This has led to an over-reliance on synthetic data for evaluation. In synthetic data, drift points are known, and thus allow for a \emph{supervised} evaluation. However, synthetic data and concept drifts are often unrealistic. It is thus not clear whether the obtained results hold true in the real-world~\cite{widmer1996learning,street2001streaming}. This, in turn, calls for an evaluation on real-world data.
\end{enumerate}
These issues give rise to the research question: \emph{How can we reliably evaluate and compare concept drift detection methods in the absence of ground truth across multiple data streams?}
We tackle this question and the limitations above with the following contributions:  
\begin{enumerate}
    \item A \textbf{drift simulation framework} that allows for controlled experiments with diverse types of drift in real-world da\-ta\-sets, enabling researchers and practitioners to better understand detector behavior in different scenarios. 
    The approach follows a Monte Carlo simulation approach where at each run predefined distribution changes are introduced into the data stream at a random point. This enables a supervised evaluation of drift detection methods while preserving the natural characteristics and inherent challenges of real-world data, addressing I2. 
    \item A set of \textbf{new drift detection evaluation metrics}, including F1 detection score, drift recall, and false alarm rate, based on a \textbf{versatile definition of correct and incorrect detections}. This enables a standardized and objective comparison of drift detection methods across different datasets and drift types, addressing I1.  
    \item A \textbf{practical standardized protocol for optimizing detectors} based on \emph{leave-one-dataset-out cross-validation}: To also address I1, we propose to optimize hyperparameters of concept drift detectors on all but one dataset and use the held-out dataset for evaluation. We hypothesize that this approach encourages the optimization process to find configurations that are more robust to distinct structural dynamics. While leave-one-group-out is a standard validation approach, to our knowledge, it has not previously been applied to drift detection methods.
\end{enumerate}

We apply the proposed framework to benchmark a set of state-of-the-art and widely used drift detection methods in a data stream classification scenario. We use 7 commonly used real-world data streams and introduce four different kinds of drift, namely class prior drift, class label swap drift, feature permutation drift, and feature filtering drift, each simulated with both abrupt and gradual transitions. 

Our results reveal that \texttt{SEED}~\cite{huang2014detecting}, \texttt{STEPD}~\cite{nishida2007detecting}, and \texttt{ABCD}~\cite{heyden2024adaptive} consistently outperform other detectors across distinct drift types, establishing new baseline benchmarks for the field. We further demonstrate that hyperparameter optimization using our proposed approach significantly improves detection performance over default configurations. 
Our implementation\footnote{\url{https://github.com/vcerqueira/experiments-drift_evaluation}} is built on the CapyMOA Python library~\cite{gomes2025capymoa}, and all code and experiment details are available for reproducibility. 

\section{Background}\label{sec:bg}


\subsection{Data Streams and Concept Drift}\label{sec:bg_cdda}

A data stream is an infinite sequence of observations generated over time according to some unknown underlying distribution. Formally, we consider a data stream $\mathcal{S}$ as a sequence of tuples $\{(X_1, y_1), (X_2, y_2), \ldots\}$, where each $X_t$ is an input (e.g., a feature\ vector) and $y_t$ is a label.\footnote{W.l.o.g., we focus on categorical labels in this work.} At time $t$, the tuple $(X_t, y_t)$ is drawn from a distribution $p_t(X, y)$, referred to as the \emph{concept}. A concept drift has occurred if this distribution changes:

\begin{definition}[Concept drift and drift point]\label{def:concept-drift}
We define a concept drift as a tuple $(d_s, d_e)$ such that $p_{d_s}(X,y) \neq p_{d_e}(X,y)$ and $d_s \leq d_e$. I.e., $d_s$ is the starting position (point of the instance where drift begins) and $d_e$ is the ending position (point where drift completes). The width of the drift is given by $w = d_e - d_s$.
\end{definition}

Concept drifts can be categorized by different types of change in the distribution. Changes in $p(X)$, termed \emph{virtual drift} (e.g.~\cite{vzliobaite2010change}), affect the input distribution but not necessarily the relationship between $X$ and $y$; such drifts can be detected without labels. Changes in $p(y|X)$, usually referred to as \emph{real drift}, alter the input-output relationship and typically require labeled data to detect. Changes may also occur in the class prior $p(y)$ or in the class-conditional feature distribution $p(X|y)$. These categories are not mutually exclusive; for instance, a shift in $p(X|y)$ 
will often also manifest as a change in $p(X)$. 

Drifts can also be categorized by how they evolve over time~\cite{gama2014survey}. In \emph{abrupt} (or sudden) drifts, the distribution shifts instantaneously from one concept to another. In \emph{gradual} drifts, two distinct concepts coexist during a transition period, with observations increasingly drawn from the new concept. In \emph{incremental} drifts, a single concept evolves continuously over time without distinct before-and-after states.
For sudden or abrupt drifts, $d_s = d_e$ and thus $w = 0$, while gradual and incremental drifts are characterized by $d_s < d_e$ and $w > 0$. While incremental drifts are relevant, our framework focuses on abrupt and gradual drifts.

\subsection{Detection Methods}\label{sec:bg_detectors}

Several methods have been proposed to detect concept drift in data streams. Most of these have been designed to track the performance (e.g.\ error rate) of a predictive model and detect whenever it worsens significantly. Such methods typically follow an approach based on either sequential analysis, control charts, or distribution monitoring. Sequential analysis approaches work by accumulating statistics over time and comparing them to a threshold. CUSUM or Page-Hinkley~\cite{page1954continuous} methods are two prominent examples following this approach. Control charts leverage the binomial distribution to define confidence intervals for the error rate, with DDM~\cite{gama2004learning} or HDDM~\cite{frias2014online} being two well-known methods. Distribution monitoring approaches work by comparing the current distribution to a reference distribution. ADWIN~\cite{bifet2007learning}, which detects significant changes in a distribution's mean, is a well-known method following this approach. We refer to the seminal survey by Gama et al.~\cite{gama2014survey} for a comprehensive review on these approaches.

In domains with considerable verification delay (the time it takes to obtain a label for a given instance), monitoring performance is impractical. In effect, other approaches track indicators that can hint to a possible performance degradation and can be monitored unsupervised. Examples of such indicators are the distribution of features (e.g.~\cite{dos2016fast,heyden2024adaptive}), the distribution of predictions~\cite{pinto2019automatic}, or an auxiliary compression loss~\cite{cerqueira2023studd}. The assumption underlying these methods is that drift in the feature space or the distribution of predictions indicate a potential performance degradation.

\subsection{Evaluating Drift Detection Methods}\label{sec:bg_rw}

Reliably evaluating and comparing concept drift detectors requires solving three interrelated sub-problems: (1) obtaining ground truth drift points in realistic data, (2) measuring performance in a way that is comparable across datasets, and (3) selecting hyperparameters without overfitting to a single stream. We organize the related work around these challenges.

\paragraph{Drift point ground truth.}
Evaluating whether a detector correctly identifies drift points requires knowing \emph{when} drifts occurred. Two paradigms have emerged in response. In \emph{proxy evaluation}, detectors are assessed indirectly by measuring whether retraining a model upon detection improves predictive performance~\cite{dos2016fast,cerqueira2023studd}. While practical, this conflates detector quality with model adaptability and does not reveal detection accuracy or timing.
\emph{Supervised evaluation} instead relies on known drift points, typically obtained from synthetic data generators that switch between predefined concepts at fixed times~\cite{bifet2017classifier,sebastiao2009study,gonccalves2014comparative}. Because ground truth is available by construction, one can directly measure detection accuracy, delay, and false alarm rates. However, synthetic streams often involve simplified distributions and unrealistic drift dynamics, leaving it unclear whether conclusions transfer to real-world settings~\cite{widmer1996learning,street2001streaming}.
No prior work provides a principled way to perform supervised evaluation on \emph{real-world} data---a gap our drift simulation framework addresses.

\paragraph{Comparability of evaluation metrics.}
Even when ground truth is available, existing metrics hinder fair cross-dataset comparison. Bifet~\cite{bifet2017classifier} proposed three widely used metrics for supervised evaluation:
\begin{itemize}
    \item Mean Time between False Alarms (MTFA): the average time span between consecutive false alarms before the true change point.
    \item Mean Detection Time (MDT): the average delay between a drift and its detection.
    \item Missed Detection Ratio (MDR): the proportion of drifts that go undetected.
\end{itemize}
\noindent These can be averaged over multiple runs for more robust estimates~\cite{faithfull2019combining,cerqueira2020unsupervised}. However, MTFA and MDT are highly dependent on stream length and drift spacing, causing their values to fluctuate across datasets and precluding meaningful cross-dataset comparison. Other works~\cite{sebastiao2009study,faithfull2019combining} have applied similar metrics under different terminology, further evidencing the lack of standardization.
To mitigate dataset dependence, Heyden et al.~\cite{heyden2024adaptive} proposed an F1 score for drift detection, defining true positives, false negatives, and false positives based on whether a drift was detected before the next change occurred. While F1 improves cross-dataset comparability, its formulation ignores the temporal aspect: a detector with unacceptable delay can still achieve perfect F1.
Our timing-aware metrics address this gap by normalizing for stream characteristics while penalizing detection delay.

\paragraph{Configuration of drift detectors.}
Beyond metrics, the hyperparameter configuration of drift detectors poses a reproducibility challenge. Most works report specific configurations without clear guidelines on how they were obtained, often relying on tacit expert knowledge or tuning on the same dataset used for evaluation~\cite{gonccalves2014comparative}. This risks overfitting to specific stream dynamics and prevents fair comparison across studies.
Our leave-one-dataset-out cross-validation protocol addresses this by optimizing hyperparameters on held-out datasets, preventing data leakage and encouraging configurations that generalize across diverse streams.

Previous benchmarks~\cite{sebastiao2009study,gonccalves2014comparative,faithfull2019combining} each suffer from one or more of the above sub-problems: they rely on synthetic data for ground truth (I2), use dataset-dependent metrics (I1), or lack a principled hyperparameter protocol (I1). Our work is the first to jointly address all three challenges within a unified framework.

\section{Drift Detection Evaluation Framework}\label{sec:eval_framework}

This section introduces a systematic framework for evaluating concept drift detection methods, including definitions of detection correctness, a set of evaluation metrics derived from these definitions, and an approach for hyperparameter tuning of detectors.


\subsection{Evaluation Criteria}\label{sec:eval_criteria}

We start by establishing clear criteria for what constitutes a correct detection, which is a key aspect for evaluating detectors in supervised settings. 
Recall our definition of concept in Definition\,\ref{def:concept-drift}: An ideal drift detector would detect data distribution changes accurately and with zero detection delay. In practice, however, a detector typically needs to first observe a certain amount of data before accurate drift detection is possible. In general, the more data a drift detector observes, the more accurate it should be. At the same time, however, this will lead to a larger detection delay. Optimizing this tradeoff is non-trivial and application-dependent.


\begin{figure}[t]
    \centering
    \includegraphics[width=.5\textwidth, trim=0cm 0cm 0cm 0cm, clip=TRUE]{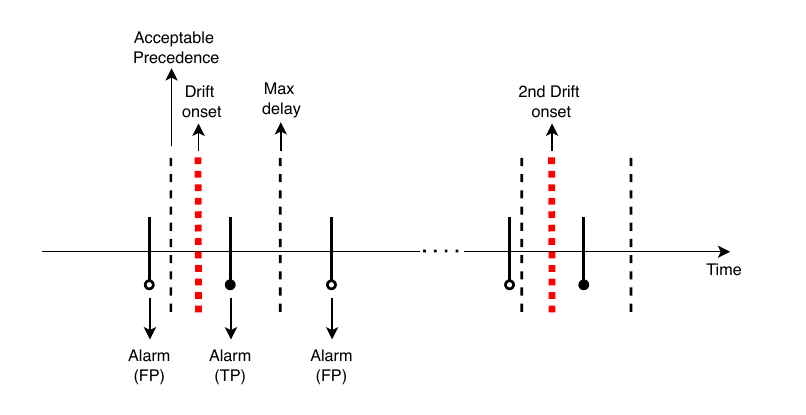}
    \caption{Drift detection evaluation criteria. The acceptable detection window is defined as the interval $[d_s - \delta_{pre}, d_e + \delta_{max}]$. Any detection falling within this window is considered a true positive, while detections outside this window are counted as false positives.}
    \label{fig:eval}
\end{figure}

To account for this trade-off, we introduce an evaluation criteria where a detection is only considered accurate if it occurs within a \textit{reasonable} time window around the actual drift point. 
Let $\delta_{max}$ denote the maximum acceptable delay parameter, representing the maximum number of instances after $d_e$ within which a detection must occur before the drift is considered missed. Similarly, let $\delta_{pre}$ denote the acceptable precedence parameter, representing the maximum number of instances before $d_s$ where an early detection can be considered valid. This precedence parameter accounts for cases where detectable changes in the data stream may precede the formal drift point. For instance, some external event may cause and precede a change in the data distribution and lead to performance degradation. While the data distribution change defines the concept drift, the preceding event is a key factor for its detection. 

We therefore define the acceptable detection window as an interval $[d_s - \delta_{pre}, d_e + \delta_{max}]$, illustrated in Figure~\ref{fig:eval}.
Any detection falling within this window is considered a true positive, while detections outside this window are counted as false positives.

For gradual drifts ($d_s < d_e$), we anchor the acceptable delay at $d_e$ rather than $d_s$. This choice ensures that the entire gradual transition falls within the acceptable window, and that $\delta_{max}$ has a consistent interpretation, i.e. maximum delay after drift completion, regardless of drift width. Note that detections occurring during the gradual transition (between $d_s$ and $d_e$) are inherently valid, as they fall within the acceptable window.

In effect, the criteria are:
\begin{itemize}
    \item \textbf{TP (True Positive)}: A concept drift occurred and was detected within the acceptable detection window $[d_s - \delta_{pre}, d_e + \delta_{max}]$.
    \item \textbf{FN (False Negative)}: A concept drift occurred but was not detected within the acceptable detection window $[d_s - \delta_{pre}, d_e + \delta_{max}]$).
    \item \textbf{FP (False Positive)}: 
    An alarm was triggered outside any acceptable detection window $[d_s - \delta_{pre}, d_e + \delta_{max}]$
\end{itemize}

\begin{remark}[True Negative]
The concept of true negative (TN) is not well defined in the context of drift detection, due to the continuity of the data stream and the fact that there are infinitely many possible moments where drift does not occur~\cite{fawcett1999activity}.
\end{remark} 

\begin{remark}[Precedence Parameter]
The value of the precedence parameter $\delta_{pre}$ is situational. For example, when introducing synthetic drift points, the distribution changes can only be detected after the drift onset, so $\delta_{pre}$ should be 0 in such cases. 
\end{remark}

\subsection{Evaluation Metrics}\label{sec:eval_metrics}

Building upon the evaluation criteria defined above, we establish several metrics to assess drift detector performance from various perspectives: accuracy, timeliness, and reliability:

\begin{itemize}
    \item \textbf{Precision}: The proportion of detected alarms that were correct ($TP/(TP+FP)$). This metric indicates how reliable the detector's alarms are. Higher precision means most alarms correspond to actual drifts rather than false positives.
    
    \item \textbf{Recall}: The proportion of actual drifts that were correctly identified ($TP/(TP+FN)$). This measure quantifies how well the detector captures actual drift events. 
    
    \item \textbf{Episode Recall}: The proportion of actual drift episodes that were correctly identified. By definition, recall counts multiple detections within the acceptable detection window as multiple TPs. Episode Recall addresses this limitation by counting multiple alarms within the same detection window as a single detection. It provides a more realistic measure of drift detection coverage.
    
    \item \textbf{F1 Score}: The harmonic mean of precision and recall.

    \item \textbf{Normalized Detection Time (NDT)}: The average number of instances between the start of a drift ($d_s$) and its detection, normalized by the maximum acceptable delay $\delta_{max}$. This metric expresses detection delay as a proportion of the acceptable window, enabling comparisons across datasets with different $\delta_{max}$ values. NDT is computed only over detected drifts (missed drifts are excluded), so values range from 0 (immediate detection) to 1 (detection at the deadline). False negatives are not covered in this metric to avoid arbitrarily large penalties (i.e. values $>>1$);
    
    \item \textbf{Alarm Rate (AR)}: The total number of alarms triggered per unit of time, calculated as $\frac{\text{Total alarms}}{\text{Total instances}} \cdot F$. By default, we use $F$ instances as the unit of time. This measures how frequently the detector signals drifts, regardless of correctness. The factor $F$ depends on the velocity of the data stream.
    
    \item \textbf{False Alarm Rate (FAR)}: The number of false positive alarms per unit of time, calculated as $\frac{\text{FP}}{\text{Total instances}} \cdot F$. This is particularly important as frequent false alarms can lead to unnecessary model updates and reduced system efficiency and trust.  
\end{itemize}

\noindent Overall, these metrics provide a complementary view on drift detector performance. For example, NDT quantifies detection latency only for successfully detected drift episodes; it does not penalize missed detections. For this reason, NDT should always be interpreted alongside Episode Recall, which captures the proportion of drift episodes that were detected within the acceptable window.

\subsection{Hyperparameter Optimization}\label{sec:hpo}

Default hyperparameters for drift detectors are often tuned w.r.t.\ the specific datasets used in the original publications.
However, using the same dataset for optimizing and evaluating the detector can cause overfitting and lead to overly optimistic performance estimates reported in the respective papers. To solve this issue, we propose a \emph{leave-one-dataset-out} cross-validation approach for optimizing the hyperparameters of drift detectors.

For each dataset held out for evaluation, the remaining datasets serve as the optimization set. A search over the detector's hyperparameter space is conducted, selecting the configuration that maximizes a chosen performance metric (e.g., F1 detection score) averaged across all scenarios in the optimization set. The detector is then evaluated on the held-out dataset using the selected configuration.

Leave-one-dataset-out cross-validation ensures that detector parameters are optimized on data different from the evaluation data, leading to more reliable performance estimates while at the same time getting the most out of detectors performance. 
The approach can be coupled with any search strategy, such as random search~\cite{bergstra2012random}.

\section{Drift Simulation Framework}\label{sec:drift_simulation}

We now introduce our method for injecting systematically controllable concept drift into real-world data.

\subsection{Simulating Drifts}

Our approach adopts a semi-synthetic methodology: given a real-world data stream, our approach injects synthetic concept drift at specified points. 

\subsubsection{Simulation Process}

Our drift simulation framework follows the prequential learning paradigm~\cite{gama2013evaluating} and consists of four steps:

\begin{enumerate}
    \item \emph{Random shuffling:} Real-world data streams, denoted $\mathcal{S}$, may already contain unknown concept drifts, which could confound the results of controlled experiments. To minimize this risk, we randomly shuffle the observations in $\mathcal{S}$, producing a permuted stream $\mathcal{S}_\pi$. 
    This step removes pre-existing temporal dependencies (e.g., seasonality) by design, bringing it closer to i.i.d., and laying the ground for controlled drift injection with known change points.

    \item \emph{Randomized drift onset:} In each prequential evaluation run, we select a single drift onset point $d_s$ at random, subject to two buffer regions. The first buffer, at the beginning of $\mathcal{S}_\pi$, ensures the classifier has sufficient training data and enough pre-drift samples for evaluating false alarms. The second buffer, at the end of the stream, provides ample post-drift samples for assessing detection performance. Once $d_s$ is established, all subsequent instances are transformed using a designated drift function (as detailed in Section~\ref{sec:drift_types}).

    \item \emph{Monte Carlo trials:} The above process (shuffling, selecting a randomized drift point, and applying drift) is repeated $k$ times. This repetition allows for robust statistical assessment of detector performance across multiple runs of the prequential workflow.

    \item \emph{Abrupt vs.\ gradual drift:} For abrupt drifts ($w = 0$), the drift transformation is applied to all instances from the drift onset onwards. For gradual drifts ($w > 0$), the transformation is applied probabilistically: the probability that an instance at time $t$ is drawn from the transformed distribution increases linearly from 0 at $d_s$ to 1 at $d_e$, computed as $p(t) = \frac{t - d_s}{w}$. After $d_e$, all instances are transformed.
\end{enumerate}

\subsubsection{Instance transformation and evaluation}

\begin{figure*}[!ht]
    \centering
    \includegraphics[width=.8\textwidth, trim=0cm 0cm 0cm 0cm, clip=TRUE]{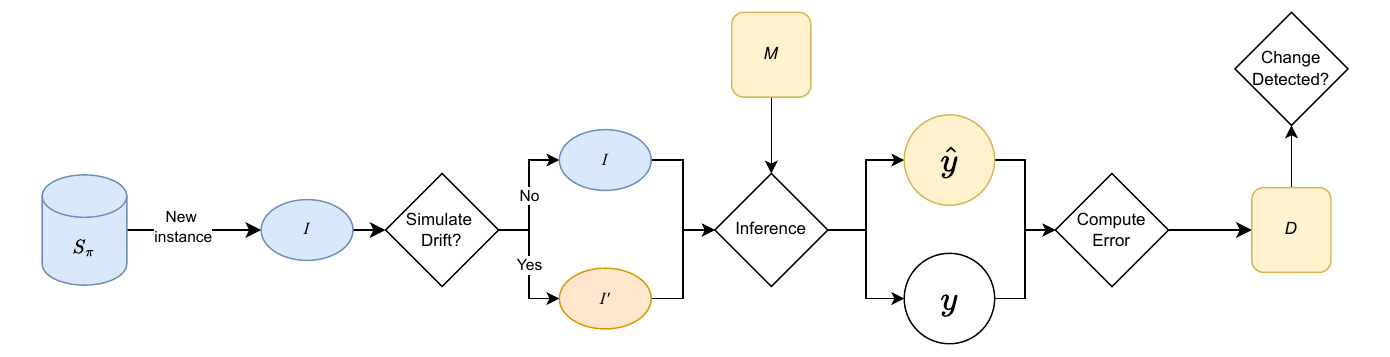}
    \caption{Inference stage of the prequential workflow with simulated drifts (using error tracking as example).}
    \label{fig:prequential_workflow_sch}
\end{figure*}

Beyond the workflow described above, the core mechanism for simulating drift is the transformation and evaluation of data instances after the drift onset. Figure~\ref{fig:prequential_workflow_sch} illustrates this process.
Let $M$ denote a predictive model (here, a classifier), $g$ the drift simulation function that applies the selected drift pattern, and $I = (X, y)$ an instance from $\mathcal{S}_\pi$.

For each incoming instance $I = (X, y)$, the framework determines whether drift should be applied based on the current position relative to the drift onset $d_s$. Before $d_s$, instances pass through unchanged. After $d_s$, the drift simulation function transforms $I$ into $I' = g(I)$ according to the specified drift pattern. The model $M$ then generates a prediction $\hat{y}$ for the (possibly transformed) instance. 
Most of the drift detectors then monitor $M$'s prediction error over time. However, other inputs are possible, such as the feature vectors~\cite{heyden2024adaptive}, or an auxiliary compression error~\cite{cerqueira2023studd}. After inference, the model $M$ is trained on the instance $I$, or the transformed instance $I'$. The complete drift simulation framework is illustrated in Algorithm~\ref{alg:drift_simulation} in appendix.

\subsection{Drift Types}\label{sec:drift_types}

The previously introduced simulation framework uses a generalized drift function $g$. This allows modeling different types of distribution changes. We now provide concrete examples of $g$, illustrating how to generate the types mentioned in Section~\ref{sec:bg_cdda}. See Appendix~\ref{app:drift-types} for the respective pseudocode.

\subsubsection{Class Prior Drift (Algorithm~\ref{alg:drift_y_prior})}

Class prior drift simulates a change in the class distribution by altering the frequency of a specific target class. After the drift onset, instances belonging to a pre-selected label $y_{sel}$ are randomly dropped from the stream with probability $p_{skip}$, effectively reducing the occurrence of that class. When an instance is dropped, the prequential workflow skips it entirely (i.e. no training or inference is performed on it), proceeding to the next observation. 
Zliobaite~\cite{vzliobaite2010change} evaluated this drift type in the context of change detection with delayed labeling, though in a non-randomized setting (e.g. fixing the proportions of the classes on specific datasets, predefined drift onset), while we create a more general framework with randomized parameters.

\subsubsection{Class Label Swap Drift (Algorithm~\ref{alg:drift_y_swap})}

Class label swap drift simulates a change in the classification function by relabeling instances of a specific class. After the drift onset, any instance originally belonging to a selected label $y_{sel}$ has its true label changed to a different target label $y_{swap}$, while its feature vector remains unchanged. This represents a scenario where the underlying classification rule changes, with the same input now corresponding to a different output. 

\subsubsection{Feature Permutation Drift (Algorithm~\ref{alg:drift_x_permute})}

Feature permutation drift introduces a structural change in the feature space by reordering the input features. After the drift onset, the features within each instance's vector $x$ are rearranged according to a fixed, randomly generated permutation $\sigma$. The class label remains unchanged. This simulates scenarios where the relative importance or ordering of features shifts. 
Zliobaite~\cite{vzliobaite2010change} also evaluated this drift in a class-conditional setting (permutation of features within a specific class), with the class and the permutation of features being hand-picked for each dataset.

\subsubsection{Feature Filtering Drift (Algorithm~\ref{alg:drift_x_exceed})}

Feature filtering drift simulates a change in the marginal distribution of a feature by conditionally filtering instances. After the drift onset, instances are checked against a threshold condition: if the value of a randomly selected numeric feature exceeds a predefined threshold $\tau$, the instance is dropped from the stream; otherwise, it passes through unchanged. When an instance is dropped, the prequential workflow skips it entirely and proceeds to the next observation. This simulates scenarios where, due to policy changes, sensor limitations, or data collection constraints, certain feature values are no longer observed. 
In this work, we consider $\tau$ to be the median of the feature values observed before the drift onset.

\subsubsection{Discussion}

The four drift types described above represent common categories of distribution change that can occur in real-world data streams. They are not intended to be exhaustive; many other forms of drift exist, such as noise injection, heteroskedasticity, or temporal dependencies~\cite{ziffer2024tenet}. Additionally, incremental drifts, where the distribution shifts continuously over an extended period~\cite{sun2025evaluation}, are also relevant but not considered in this work. Our selection focuses on distinct, interpretable drift patterns that allow for controlled experimentation.

Each drift type involves random parameter choices (e.g., which class to affect, which feature to filter). This variability means that a single run may not fully characterize detector performance. The Monte Carlo approach described earlier addresses this by repeating the evaluation across multiple runs with different random configurations, leading to more robust performance estimates. For simplicity, each Monte Carlo run contains a single drift, whereas real-world streams may exhibit multiple ones. These scenarios can be framed as different episodes of drift events as illustrated in Figure~\ref{fig:eval}.

Finally, because each drift type transforms instances from $I$ to $I'$ (or filters them), all four can be applied in either abrupt or gradual mode as described in Section~\ref{sec:drift_simulation}. This flexibility allows the framework to simulate a wide range of drift dynamics.

\section{Experiments}\label{sec:experiments}

\subsection{Experimental Setup}

\paragraph{Datasets}

We evaluate our approach on seven widely used real-world datasets from the USP Data Stream Repository~\cite{souza2020challenges}: Asfault~\cite{souza2018asphalt}, Electricity~\cite{harries1999splice}, Covertype~\cite{blackard1999comparative}, GasSensorArray~\cite{vergara2012chemical}, NOAA~\cite{ditzler2012incremental}, Posture~\cite{dos2016fast}, and Rialto~\cite{losing2016knn}. Detailed descriptions of each dataset are available in the repository\footnote{\url{https://sites.google.com/view/uspdsrepository}}.

To create controlled evaluation settings, we systematically generate semi-synthetic data streams by injecting known concept drift points into these datasets using the proposed drift simulation framework. For each dataset and drift type, we perform 50 Monte Carlo trials. In each trial, the drift onset time is randomly sampled between 50\% and 80\% of the total dataset length. Drift width and maximum delay parameters are dataset-specific. These were selected based on dataset length and are summarized in Table~\ref{tab:drift_parameters}. The precedence parameter is fixed at zero ($\delta_{pre} = 0$) for all datasets since synthetic drifts cannot be not anticipated before their onset.




We benchmark a total of 14 change detection methods: \texttt{ADWIN}~\cite{bifet2007learning}, \texttt{CUSUM}~\cite{page1954continuous}, \texttt{DDM}~\cite{gama2004learning}, \texttt{EDDM}~\cite{baena2006early}, \texttt{EWMA}~\cite{ross2012exponentially}, \texttt{GMA}~\cite{roberts2000control}, \texttt{HDDMA}~\cite{frias2014online}, \texttt{HDDMW}~\cite{frias2014online}, \texttt{PH}~\cite{page1954continuous}, \texttt{RDDM}~\cite{barros2017rddm}, \texttt{SEED}~\cite{huang2014detecting}, \texttt{STEPD}~\cite{nishida2007detecting}, \texttt{ABCD} and \texttt{ABCD(X)}~\cite{heyden2024adaptive}, and \texttt{STUDD}~\cite{cerqueira2023studd}. Most of these are based on tracking the performance of a classifier, namely the instance-wise error rate, except for \texttt{ABCD(X)}~\cite{heyden2024adaptive}, which monitors the feature space and \texttt{STUDD}, which monitors an auxiliary compression loss~\cite{cerqueira2023studd}. Most of these methods follow a distribution monitoring, control chart, or sequential analysis approach. The exception is \texttt{STUDD}, which, being a meta-detector, can be coupled with any of the other methods. In this work, \texttt{STUDD} is coupled with \texttt{ADWIN}. We also note that \texttt{ABCD} can be applied both to the error rate and the feature space. We denote the latter as \texttt{ABCD(X)}. 
We note that our main objective is to establish a standardized evaluation protocol rather than to identify the best performing detector. Thus, we focus on well-established detection methods whose performance characteristics are already well understood.

We select the Hoeffding Tree~\cite{domingos2000mining} as the classifier in the experiments, a well-established algorithm in the streaming machine learning literature. While ensembles of Hoeffding Trees (e.g.~\cite{gomes2017adaptive}) have been shown to be more accurate, we focus on the Hoeffding Tree due to its computational efficiency. We use default hyperparameters on Hoeffding Tree classifier across all experiments\footnote{\url{https://capymoa.org/api/modules/capymoa.classifier.HoeffdingTree.html}}.

\paragraph{Hyperparameter Optimization.}

Regarding hyperparameter optimization, we conduct the leave-one-dataset-out approach described in Section~\ref{sec:hpo}. When evaluating the detectors on each dataset, we use the remaining datasets for optimizing its hyperparameters based on the F1 detection score. The optimization is conducted using 30 iterations of random search. Each configuration is also evaluted using the Monte Carlo procedure described in Section~\ref{sec:hpo}. The configuration space for each detector is presented in Table~\ref{tab:hpo_config} in appendix.

\subsection{Results}\label{sec:results}

Most of the results presented below are based on F1 score. The scores on the remaining metrics (including execution time), other classifiers and synthetic data streams, are available in the Appendix~\ref{app:results}. In all tables, bold (underlined) scores denote the best (second-best) detector in the respective scenario. 

\subsubsection{Overall scores.} 

Tables~\ref{tab:avgrank_abr} and~\ref{tab:avgrank_grd} show the average rank based on F1 detection score for each drift detection method across different drift scenarios, for abrupt and gradual drifts respectively. The average rank of a detector denotes its relative position according to performance; lower values indicate better performance.

\begin{table}[t]
\caption{Average rank of drift detectors across different datasets for abrupt drifts}
\label{tab:avgrank_abr}
\resizebox{.45\textwidth}{!}{%
\begin{tabular}{l>{\raggedleft\arraybackslash}p{1.3cm}>{\raggedleft\arraybackslash}p{1.3cm}>{\raggedleft\arraybackslash}p{1.3cm}>{\raggedleft\arraybackslash}p{1.3cm}}
\toprule
Detector & Feature Filtering & Feat. Permutation & Class Prior & Class Swap \\
\midrule
\texttt{ABCD} & 8.1 & 3.9 & 8.0 & 3.6 \\
\texttt{ABCD(X)} & 12.7 & \textbf{1.0} & 13.1 & 13.1 \\
\texttt{ADWIN} & \underline{4.1} & 7.6 & \underline{4.1} & 7.1 \\
\texttt{CUSUM} & 5.6 & 11.9 & 6.1 & 9.2 \\
\texttt{DDM} & 7.3 & 11.4 & 6.0 & 9.9 \\
\texttt{EWMA} & 9.7 & 12.9 & 9.4 & 10.9 \\
\texttt{GMA} & 9.0 & 7.7 & 8.7 & 7.3 \\
\texttt{HDDMA} & 7.4 & 5.6 & 6.3 & 5.4 \\
\texttt{HDDMW} & 7.1 & 6.9 & 5.8 & 6.0 \\
\texttt{PH} & 13.1 & 12.9 & 11.6 & 10.1 \\
\texttt{RDDM} & 6.6 & 10.2 & 7.9 & 9.1 \\
\texttt{SEED} & 4.4 & \underline{2.7} & \textbf{3.9} & \texttt{1.1} \\
\texttt{STEPD} & \textbf{3.4} & 3.7 & 4.3 & \underline{2.4} \\
\texttt{STUDD} & 6.4 & 6.8 & 9.9 & 9.6 \\
\bottomrule
\end{tabular}%
}
\end{table}

The results reveal distinct tiers of detector performance. \texttt{SEED} and \texttt{STEPD} exhibit consistently competitive F1 scores across all drift types and abruptness conditions, ranking among the top three performers in most scenarios.

Several detectors perform well under specific conditions but struggle in others. Among performance-tracking methods, \texttt{ADWIN} excels at feature filtering and class prior drifts but underperforms on feature permutation. \texttt{ABCD} achieves top-tier ranks on feature permutation and class swaps while showing weaker performance on feature filtering. \texttt{DDM} shows moderate results overall, performing best on class prior under gradual conditions. \texttt{GMA} demonstrates particular strength on feature filtering in gradual settings, achieving the best average rank for that scenario. \texttt{HDDMW} shows stable middle-tier performance (average ranks around 5 to 7) across most conditions.

The unsupervised detectors, \texttt{ABCD(X)} and \texttt{STUDD}, exhibit a characteristic pattern: reasonable or strong performance on feature-space drifts but poor results on label-based changes. \texttt{ABCD(X)} achieves near-perfect relative detection of feature permutation drifts (rank 1.0 for abrupt, 3.3 for gradual) but fails on class prior and class swap drifts. \texttt{STUDD} ranks second on feature filtering for gradual drifts but is less effective for label-based scenarios. This behavior is expected, as these methods cannot observe changes that only affect the target variable.
Finally, \texttt{PH} and \texttt{EWMA} consistently rank among the worst performers across nearly all scenarios, with average ranks typically exceeding 10.

\begin{table}[th]
\caption{Average rank of drift detectors across different datasets for gradual drifts}
\label{tab:avgrank_grd}
\resizebox{.45\textwidth}{!}{%
\begin{tabular}{l>{\raggedleft\arraybackslash}p{1.3cm}>{\raggedleft\arraybackslash}p{1.3cm}>{\raggedleft\arraybackslash}p{1.3cm}>{\raggedleft\arraybackslash}p{1.3cm}}
\toprule
Detector & Feature Filtering & Feat. Permutation & Class Prior & Class Swap \\
\midrule
\texttt{ABCD} & 11.7 & 4.1 & 9.8 & 5.1 \\
\texttt{ABCD(X)} & 13.0 & 3.3 & 13.0 & 13.1 \\
\texttt{ADWIN} & 6.0 & 11.2 & 6.8 & 7.9 \\
\texttt{CUSUM} & 6.3 & 11.9 & 7.3 & 9.7 \\
\texttt{DDM} & 6.1 & 9.4 & \textbf{4.7} & 8.1 \\
\texttt{EWMA} & 11.7 & 12.4 & 10.6 & 12.0 \\
\texttt{GMA} & \textbf{3.9} & 6.6 & 6.1 & 4.7 \\
\texttt{HDDMA} & 9.7 & 7.9 & 7.8 & 10.5 \\
\texttt{HDDMW} & 5.5 & 6.3 & 5.2 & 5.4 \\
\texttt{PH} & 10.9 & 12.4 & 9.7 & 11.1 \\
\texttt{RDDM} & 4.7 & 6.4 & 6.6 & 5.3 \\
\texttt{SEED} & 5.9 & \textbf{2.7} & \underline{5.0} & \texttt{1.9} \\
\texttt{STEPD} & 5.4 & \underline{2.8} & 5.2 & \underline{2.0} \\
\texttt{STUDD} & \underline{4.2} & 7.6 & 7.1 & 8.1 \\
\bottomrule
\end{tabular}%
}
\end{table}

\subsubsection{Abrupt vs Gradual.} 

\begin{figure}[!ht]
    \centering
    \includegraphics[width=.5\textwidth, trim=0cm 0cm 0cm 0cm, clip=TRUE]{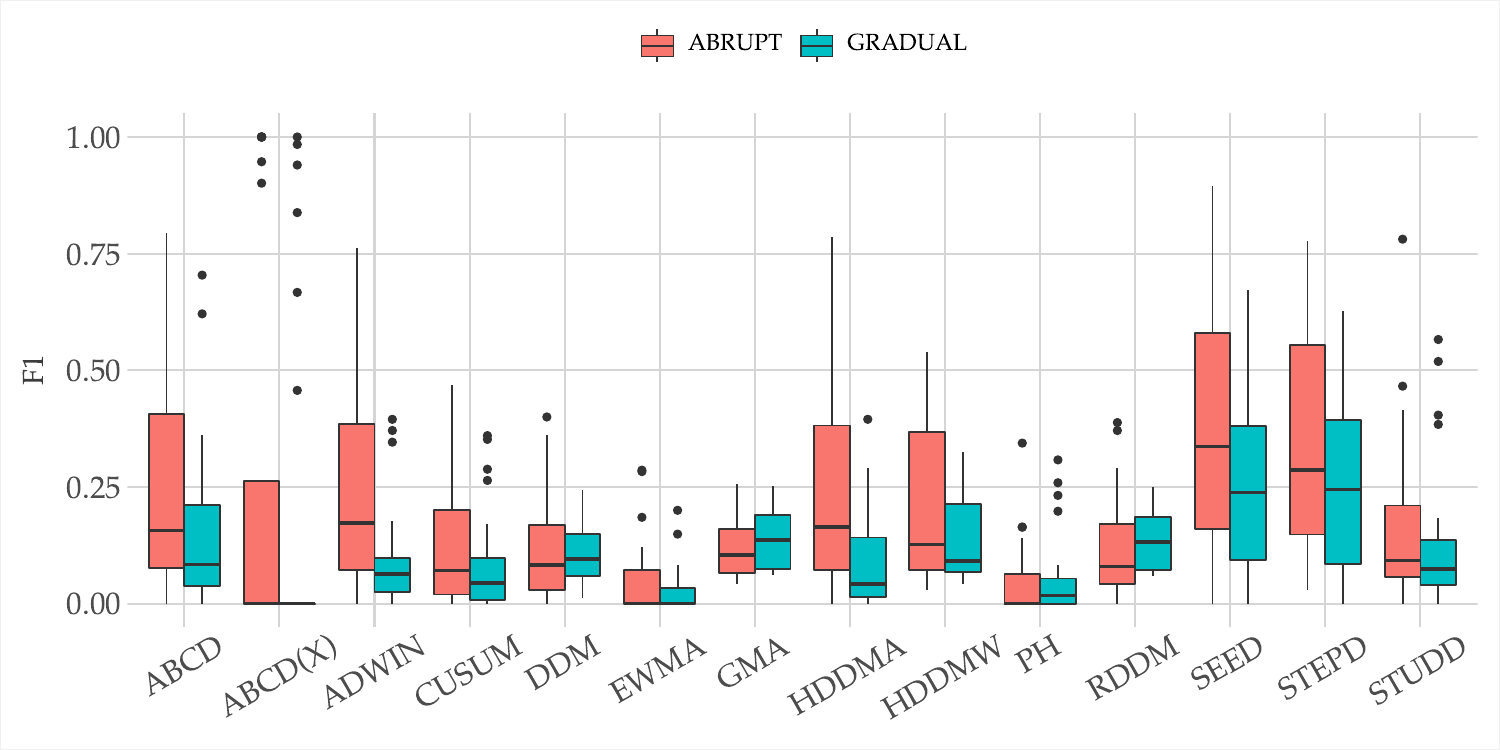}
    \caption{Distribution of F1 detection scores across dataset/drift-type pairs for each detector, comparing abrupt and gradual drift conditions.}
    \label{fig:plot4_abrupt_gradual}
    \Description{Boxplot comparing F1 scores of 14 drift detectors under abrupt and gradual drift conditions.}
\end{figure}

Figure~\ref{fig:plot4_abrupt_gradual} compares abrupt versus gradual scenarios by showing the F1 score distribution for each detector. Each boxplot aggregates $4 \times 7 = 28$ data points (4 drift types $\times$ 7 datasets).

As expected, abrupt drifts are generally easier to detect: most detectors exhibit higher median F1 scores and greater upward variance for abrupt conditions. \texttt{SEED} and \texttt{STEPD} maintain the highest medians in both conditions, though with notable performance degradation under gradual drifts. 

\subsubsection{F1 vs FAR} 

\begin{figure}[htb]
    \centering
    \includegraphics[width=.4\textwidth, trim=0cm 0cm 0cm 0cm, clip=TRUE]{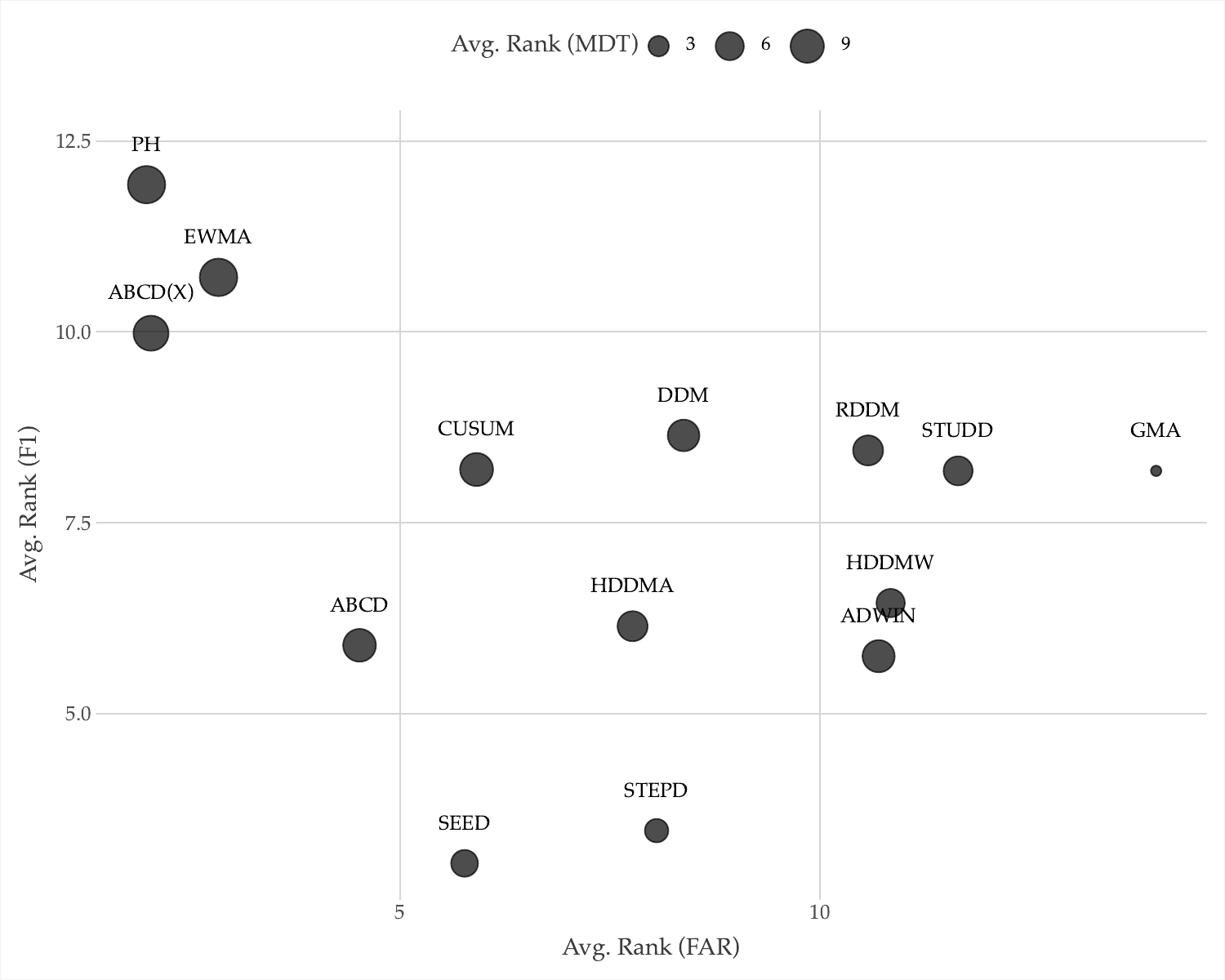}
    \caption{Trade-off between F1 average rank (y-axis, lower is better) and FAR average rank (x-axis, lower is better). Dot size encodes MDT average rank (smaller is faster detection).}
    \label{fig:plot7_f1_far_mdt}
    \Description{Scatterplot showing F1 vs FAR trade-off for 14 drift detectors, with MDT encoded as dot size.}
\end{figure}

Figure~\ref{fig:plot7_f1_far_mdt} visualizes the trade-off between detection accuracy (F1 average rank) and reliability (FAR average rank), with detection timing (MDT average rank) encoded as dot size. Lower values indicate better performance on all three axes; an ideal detector would appear in the bottom-left corner with a small dot.

\texttt{SEED} achieves the best F1 rank while maintaining a moderate FAR, making it the most balanced choice overall. \texttt{STEPD} follows closely with similar characteristics. \texttt{ABCD} offers a compelling alternative when minimizing false alarms is a priority; it achieves the best FAR rank among detectors with competitive F1 scores, though with slightly lower F1 performance than \texttt{SEED} and \texttt{STEPD}.


\subsubsection{Detector hyperparameter optimization.} 

\begin{figure}[htb]
    \centering
    \includegraphics[width=.5\textwidth, trim=0cm 0cm 0cm 0cm, clip=TRUE]{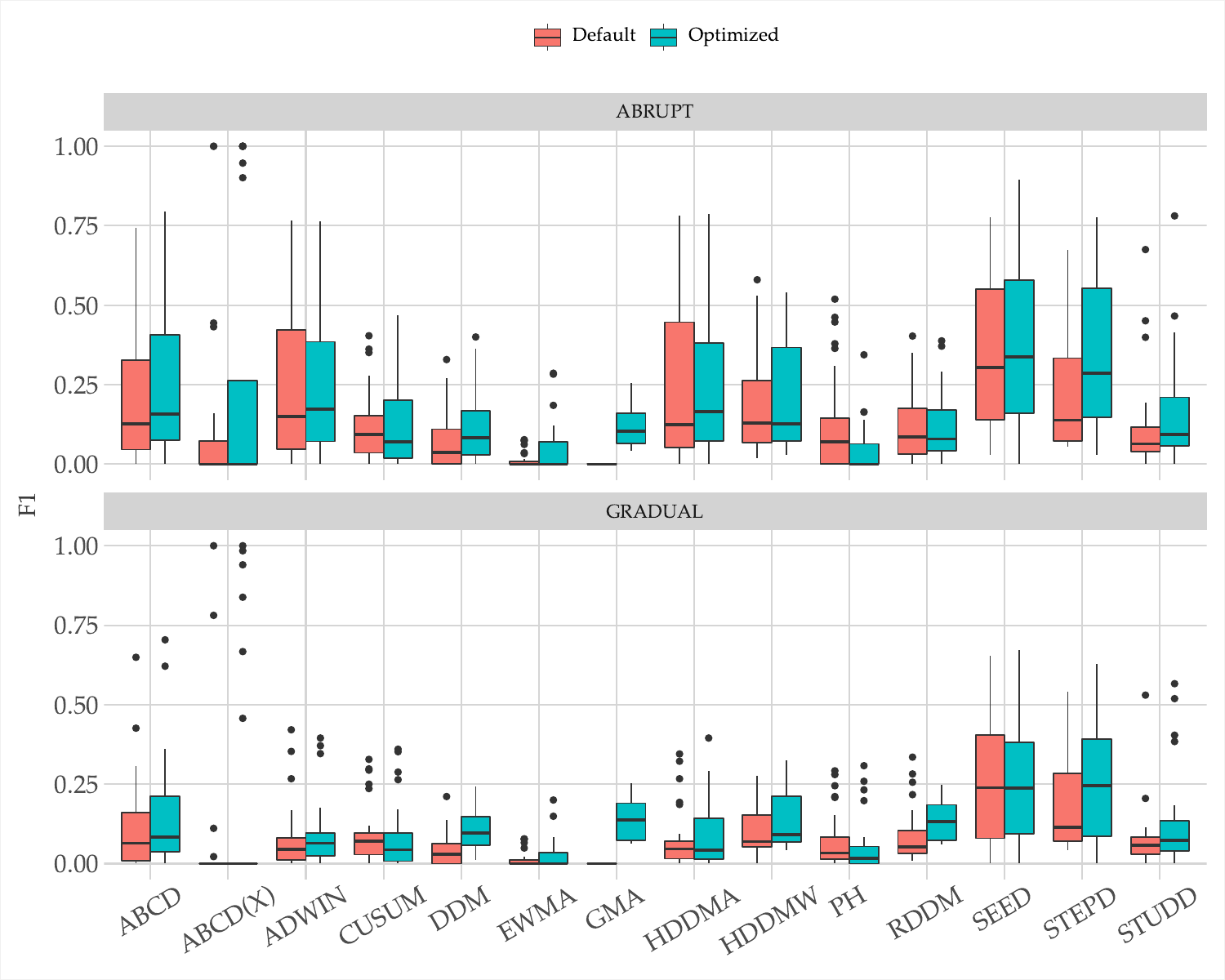}
    \caption{Distribution of F1 scores across different dataset/drift type pairs for each detection method controlling for hyperparameter optimization approach and drift abruptness.}
    \label{fig:plot1_hypertuning}
\end{figure}

Finally, we analyse the impact of the hyperparameter optimization described in Section~\ref{sec:hpo}. The results reported so far show the scores of each detector after applying hyperparameter optimization. Figure~\ref{fig:plot1_hypertuning} shows the distribution of F1 scores across different dataset/drift type pairs for each detection method controlling for hyperparameter optimization approach: optimized and default parameters. The results show that most approaches improve their median F1 score when optimized, apart from \texttt{SEED}.

\subsection{Discussion}\label{sec:discussion}

\paragraph{Main findings.}

The experiments reveal distinct patterns in detector performance. We highlight the following main findings:
\begin{enumerate}
    
    \item \texttt{SEED} and \texttt{STEPD} exhibit the most consistent performance across all drift types and abruptness conditions, ranking among the top three in most scenarios. When minimizing false alarms is a priority, \texttt{ABCD} offers a compelling alternative with the best FAR rank among accurate detectors, though with slightly lower F1 than \texttt{SEED} and \texttt{STEPD}.

    \item Gradual drifts are systematically harder to detect than abrupt ones. Most detectors show lower median F1 scores and reduced variance under gradual conditions, with the performance gap particularly pronounced for methods like \texttt{ABCD} and \texttt{ABCD(X)}.

    \item Unsupervised detectors (\texttt{ABCD(X)} and \texttt{STUDD}) perform better on feature-space drifts (feature permutation, feature filtering) than on label-based changes (class prior, class swaps), as expected since they cannot observe target variable shifts.

    \item Hyperparameter optimization yields improvements for most detectors, with gains more pronounced for abrupt drifts. Notably, \texttt{STEPD}'s strong performance is largely attributable to effective tuning. \texttt{PH} and \texttt{EWMA} remain ineffective regardless of configuration, suggesting fundamental limitations rather than suboptimal defaults.
\end{enumerate}

These findings offer practical guidance for practitioners: \texttt{SEED} and \texttt{STEPD} are robust defaults when drift characteristics are unknown, while specialized detectors may be preferred when the drift type or false alarm tolerance is constrained.

\paragraph{Limitations and future work.}

While the experiments provide valuable insights, there are some limitations and potential future work. First, the experiments are limited to one classifier (Hoeffding tree) and four drift types simulated in 7 real-world datasets. Nonetheless, the proposed drift simulation framework is generic and can be extended with additional drift definitions, or mixtures of them. Another limitation is that the prequential workflow is simplified in the sense that it assumes immediate feedback, with labels being readily available at each step after inference. In practice, this is often not the case~\cite{cerqueira2023studd}. 
Finally, in our drift simulation procedure, we randomly shuffle the original data stream to eliminate any pre-existing drifts that might confound the evaluation. This approach preserves the real-world feature-level complexity of the data; however, it also removes any inherent temporal structure. Preserving temporal dependencies in future extensions of the framework would enable a more comprehensive and realistic assessment of detector performance.


\section{Conclusion}\label{sec:conclusion}

This work contributes with a unified framework for evaluating and comparing change detection methods in data streams. The novelty of the framework settles on three main contributions: (1) a flexible drift simulation framework that allows for controlled experiments with various types of drift, enabling practitioners to better understand detector behavior under different drift conditions; (2) a systematic evaluation criteria to compare drift detection methods, including new evaluation metrics based on detection precision and recall, with the goal of building a common ground for the evaluation of drift detection methods; (3) an approach for hyperparameter optimization of drift detection methods, based on a leave-one-dataset-out scheme. 

We applied the proposed framework to benchmark state-of-the-art and widely used drift detection methods on synthetic and real-world datasets. The experiments reveal interesting patterns in the performance of the detectors. While there is a noticeable variability in relative performance across different conditions, we highlight that \texttt{SEED}, \texttt{STEPD} and \texttt{ABCD} show the best overall performance across all datasets and drift types. 

\section*{Acknowledgements}

This work is funded by national funds through FCT – Foundation for Science and Technology, I.P., within the scope of the research unit UID/00326 - Centre for Informatics and Systems of the University of Coimbra, \url{https://doi.org/10.54499/UID/00326/2025}. Heitor Murilo Gomes acknowledges the financial support of the Marsden Fund under award number VUW2213.

\bibliographystyle{ACM-Reference-Format}
\bibliography{references}

@inproceedings{fawcett1999activity,
  title={Activity monitoring: Noticing interesting changes in behavior},
  author={Fawcett, Tom and Provost, Foster},
  booktitle={Proceedings of the fifth ACM SIGKDD international conference on Knowledge discovery and data mining},
  pages={53--62},
  year={1999}
}

@inproceedings{sebastiao2009study,
  title={A study on change detection methods},
  author={Sebastiao, Raquel and Gama, Joao},
  booktitle={Progress in artificial intelligence, 14th Portuguese conference on artificial intelligence, EPIA},
  volume={2009},
  year={2009}
}

@article{faithfull2019combining,
  title={Combining univariate approaches for ensemble change detection in multivariate data},
  author={Faithfull, William J and Rodr{\'\i}guez, Juan J and Kuncheva, Ludmila I},
  journal={Information Fusion},
  volume={45},
  pages={202--214},
  year={2019},
  publisher={Elsevier}
}

@article{gonccalves2014comparative,
  title={A comparative study on concept drift detectors},
  author={Gon{\c{c}}alves Jr, Paulo M and de Carvalho Santos, Silas GT and Barros, Roberto SM and Vieira, Davi CL},
  journal={Expert Systems with Applications},
  volume={41},
  number={18},
  pages={8144--8156},
  year={2014},
  publisher={Elsevier}
}

@article{gomes2025capymoa,
  title={CapyMOA: efficient machine learning for data streams in python},
  author={Gomes, Heitor Murilo and Lee, Anton and Gunasekara, Nuwan and Sun, Yibin and Cassales, Guilherme Weigert and Liu, Justin and Heyden, Marco and Cerqueira, Vitor and Bahri, Maroua and Koh, Yun Sing and others},
  journal={arXiv preprint arXiv:2502.07432},
  year={2025}
}

@inproceedings{nishida2007detecting,
  title={Detecting concept drift using statistical testing},
  author={Nishida, Kyosuke and Yamauchi, Koichiro},
  booktitle={International conference on discovery science},
  pages={264--269},
  year={2007},
  organization={Springer}
}

@inproceedings{huang2014detecting,
  title={Detecting volatility shift in data streams},
  author={Huang, David Tse Jung and Koh, Yun Sing and Dobbie, Gillian and Pears, Russel},
  booktitle={2014 IEEE International Conference on Data Mining},
  pages={863--868},
  year={2014},
  organization={IEEE}
}

@article{barros2017rddm,
  title={RDDM: Reactive drift detection method},
  author={Barros, Roberto SM and Cabral, Danilo RL and Gon{\c{c}}alves Jr, Paulo M and Santos, Silas GTC},
  journal={Expert Systems with Applications},
  volume={90},
  pages={344--355},
  year={2017},
  publisher={Elsevier}
}

@article{roberts2000control,
  title={Control chart tests based on geometric moving averages},
  author={Roberts, Stuart W},
  journal={Technometrics},
  volume={42},
  number={1},
  pages={97--101},
  year={2000},
  publisher={Taylor \& Francis}
}

@article{souza2020challenges,
     title={Challenges in Benchmarking Stream Learning Algorithms with Real-world Data},
     author={Souza, V. M. A. and Reis, D. M. and Maletzke, A. G. and Batista, G. E. A. P. A.},
     journal={Data Mining and Knowledge Discovery},
     pages={1805-1858},
     volume={34},
     year={2020},
     doi={10.1007/s10618-020-00698-5}
}

@inproceedings{losing2016knn,
  title={KNN classifier with self adjusting memory for heterogeneous concept drift},
  author={Losing, Viktor and Hammer, Barbara and Wersing, Heiko},
  booktitle={2016 IEEE 16th international conference on data mining (ICDM)},
  pages={291--300},
  year={2016},
  organization={IEEE}
}

@article{bergstra2012random,
  title={Random search for hyper-parameter optimization},
  author={Bergstra, James and Bengio, Yoshua},
  journal={The journal of machine learning research},
  volume={13},
  number={1},
  pages={281--305},
  year={2012},
  publisher={JMLR. org}
}

@article{ditzler2012incremental,
  title={Incremental learning of concept drift from streaming imbalanced data},
  author={Ditzler, Gregory and Polikar, Robi},
  journal={IEEE transactions on knowledge and data engineering},
  volume={25},
  number={10},
  pages={2283--2301},
  year={2012},
  publisher={IEEE}
}

@article{vergara2012chemical,
  title={Chemical gas sensor drift compensation using classifier ensembles},
  author={Vergara, Alexander and Vembu, Shankar and Ayhan, Tuba and Ryan, Margaret A and Homer, Margie L and Huerta, Ram{\'o}n},
  journal={Sensors and Actuators B: Chemical},
  volume={166},
  pages={320--329},
  year={2012},
  publisher={Elsevier}
}

@article{souza2018asphalt,
  title={Asphalt pavement classification using smartphone accelerometer and complexity invariant distance},
  author={Souza, Vinicius MA},
  journal={Engineering Applications of Artificial Intelligence},
  volume={74},
  pages={198--211},
  year={2018},
  publisher={Elsevier}
}

@article{ross2012exponentially,
  title={Exponentially weighted moving average charts for detecting concept drift},
  author={Ross, Gordon J and Adams, Niall M and Tasoulis, Dimitris K and Hand, David J},
  journal={Pattern recognition letters},
  volume={33},
  number={2},
  pages={191--198},
  year={2012},
  publisher={Elsevier}
}

@inproceedings{baena2006early,
  title={Early drift detection method},
  author={Baena-Garc{\i}a, Manuel and del Campo-{\'A}vila, Jos{\'e} and Fidalgo, Raul and Bifet, Albert and Gavalda, Ricard and Morales-Bueno, Rafael},
  booktitle={Fourth international workshop on knowledge discovery from data streams},
  volume={6},
  pages={77--86},
  year={2006}
}

@article{gomes2017adaptive,
  title={Adaptive random forests for evolving data stream classification},
  author={Gomes, Heitor M and Bifet, Albert and Read, Jesse and Barddal, Jean Paul and Enembreck, Fabr{\'\i}cio and Pfharinger, Bernhard and Holmes, Geoff and Abdessalem, Talel},
  journal={Machine Learning},
  volume={106},
  number={9},
  pages={1469--1495},
  year={2017},
  publisher={Springer}
}

@inproceedings{domingos2000mining,
  title={Mining high-speed data streams},
  author={Domingos, Pedro and Hulten, Geoff},
  booktitle={Proceedings of the sixth ACM SIGKDD international conference on Knowledge discovery and data mining},
  pages={71--80},
  year={2000}
}

@article{harries1999splice,
  title={Splice-2 comparative evaluation: Electricity pricing},
  author={Harries, Michael and Wales, New South and others},
  year={1999},
  publisher={University of New South Wales, School of Computer Science and Engineering}
}

@article{blackard1999comparative,
  title={Comparative accuracies of artificial neural networks and discriminant analysis in predicting forest cover types from cartographic variables},
  author={Blackard, Jock A and Dean, Denis J},
  journal={Computers and electronics in agriculture},
  volume={24},
  number={3},
  pages={131--151},
  year={1999},
  publisher={Elsevier}
}

@article{gama2013evaluating,
  title={On evaluating stream learning algorithms},
  author={Gama, Joao and Sebastiao, Raquel and Rodrigues, Pedro Pereira},
  journal={Machine learning},
  volume={90},
  number={3},
  pages={317--346},
  year={2013},
  publisher={Springer}
}

@inproceedings{street2001streaming,
  title={A streaming ensemble algorithm (SEA) for large-scale classification},
  author={Street, W Nick and Kim, YongSeog},
  booktitle={Proceedings of the seventh ACM SIGKDD international conference on Knowledge discovery and data mining},
  pages={377--382},
  year={2001}
}

@article{widmer1996learning,
  title={Learning in the presence of concept drift and hidden contexts},
  author={Widmer, Gerhard and Kubat, Miroslav},
  journal={Machine learning},
  volume={23},
  number={1},
  pages={69--101},
  year={1996},
  publisher={Springer}
}

@inproceedings{vzliobaite2010change,
  title={Change with delayed labeling: When is it detectable?},
  author={{\v{Z}}liobaite, Indre},
  booktitle={2010 IEEE international conference on data mining workshops},
  pages={843--850},
  year={2010},
  organization={IEEE}
}

@inproceedings{bifet2007learning,
  title={Learning from time-changing data with adaptive windowing},
  author={Bifet, Albert and Gavalda, Ricard},
  booktitle={Proceedings of the 2007 SIAM international conference on data mining},
  pages={443--448},
  year={2007},
  organization={SIAM}
}

@inproceedings{dos2016fast,
  title={Fast unsupervised online drift detection using incremental kolmogorov-smirnov test},
  author={Dos Reis, Denis Moreira and Flach, Peter and Matwin, Stan and Batista, Gustavo},
  booktitle={Proceedings of the 22nd ACM SIGKDD international conference on knowledge discovery and data mining},
  pages={1545--1554},
  year={2016}
}

@article{cerqueira2023studd,
  title={STUDD: A student--teacher method for unsupervised concept drift detection},
  author={Cerqueira, Vitor and Gomes, Heitor Murilo and Bifet, Albert and Torgo, Luis},
  journal={Machine Learning},
  volume={112},
  number={11},
  pages={4351--4378},
  year={2023},
  publisher={Springer}
}

@article{pinto2019automatic,
  title={Automatic model monitoring for data streams},
  author={Pinto, F{\'a}bio and Sampaio, Marco OP and Bizarro, Pedro},
  journal={arXiv preprint arXiv:1908.04240},
  year={2019}
}

@article{heyden2024adaptive,
  title={Adaptive Bernstein change detector for high-dimensional data streams},
  author={Heyden, Marco and Fouch{\'e}, Edouard and Arzamasov, Vadim and Fenn, Tanja and Kalinke, Florian and B{\"o}hm, Klemens},
  journal={Data Mining and Knowledge Discovery},
  volume={38},
  number={3},
  pages={1334--1363},
  year={2024},
  publisher={Springer}
}

@article{gama2014survey,
  title={A survey on concept drift adaptation},
  author={Gama, Jo{\~a}o and {\v{Z}}liobait{\.e}, Indr{\.e} and Bifet, Albert and Pechenizkiy, Mykola and Bouchachia, Abdelhamid},
  journal={ACM computing surveys (CSUR)},
  volume={46},
  number={4},
  pages={1--37},
  year={2014},
  publisher={ACM New York, NY, USA}
}

@inproceedings{gama2004learning,
  title={Learning with drift detection},
  author={Gama, Joao and Medas, Pedro and Castillo, Gladys and Rodrigues, Pedro},
  booktitle={Brazilian symposium on artificial intelligence},
  pages={286--295},
  year={2004},
  organization={Springer}
}

@inproceedings{cerqueira2020unsupervised,
  title={Unsupervised concept drift detection using a student--teacher approach},
  author={Cerqueira, Vitor and Gomes, Heitor Murilo and Bifet, Albert},
  booktitle={International conference on discovery science},
  pages={190--204},
  year={2020},
  organization={Springer}
}

@inproceedings{bifet2017classifier,
  title={Classifier concept drift detection and the illusion of progress},
  author={Bifet, Albert},
  booktitle={International conference on artificial intelligence and soft computing},
  pages={715--725},
  year={2017},
  organization={Springer}
}

@article{frias2014online,
  title={Online and non-parametric drift detection methods based on Hoeffding’s bounds},
  author={Frias-Blanco, Isvani and del Campo-{\'A}vila, Jos{\'e} and Ramos-Jimenez, Gonzalo and Morales-Bueno, Rafael and Ortiz-Diaz, Agustin and Caballero-Mota, Yail{\'e}},
  journal={IEEE Transactions on Knowledge and Data Engineering},
  volume={27},
  number={3},
  pages={810--823},
  year={2014},
  publisher={IEEE}
}

@article{page1954continuous,
  title={Continuous inspection schemes},
  author={Page, Ewan S},
  journal={Biometrika},
  volume={41},
  number={1/2},
  pages={100--115},
  year={1954},
  publisher={JSTOR}
}

@article{sun2025evaluation,
  title={Evaluation for Regression Analyses on Evolving Data Streams},
  author={Sun, Yibin and Gomes, Heitor Murilo and Pfahringer, Bernhard and Bifet, Albert},
  journal={arXiv preprint arXiv:2502.07213},
  year={2025}
}

@inproceedings{ziffer2024tenet,
  title={Tenet: Benchmarking Data Stream Classifiers in Presence of Temporal Dependence},
  author={Ziffer, Giacomo and Giannini, Federico and Della Valle, Emanuele},
  booktitle={2024 IEEE International Conference on Big Data (BigData)},
  pages={1187--1196},
  year={2024},
  organization={IEEE}
}

\appendix

\section{Drift Simulation}

\subsection{Simulation Framework}

Algorithm~\ref{alg:drift_simulation} describes the procedure for simulating drift on real-world data streams based on Monte Carlo trials.

\begin{algorithm}[htb]
    \caption{Drift Simulation Framework}
    \label{alg:drift_simulation}
    \begin{algorithmic}[1]
    \Require Data stream $\mathcal{S}$, Drift function $g$, Predictive model $M$, Change detector $D$
    \Require Number of trials $k$, Drift width $w$, Buffer sizes $b_{start}$, $b_{end}$
    \Ensure Detection results for each trial
    \For{$j = 1$ to $k$} \Comment{Monte Carlo trials}
        \State $\mathcal{S}_\pi \gets \textsc{Shuffle}(\mathcal{S})$ \Comment{Random permutation}
        \State $d_s \gets \textsc{RandomInt}(b_{start}, |\mathcal{S}_\pi| - b_{end})$ \Comment{Random drift onset}
        \State $d_e \gets d_s + w$ \Comment{Drift end position}
        \State \textsc{Reset}($M$, $D$) \Comment{Initialize model and detector}
        \For{$i = 1$ to $|\mathcal{S}_\pi|$}
            \State $I \gets \mathcal{S}_\pi[i]$ \Comment{Get instance $(X, y)$}
            \If{$i < d_s$} \Comment{Before drift onset}
                \State $I' \gets I$
            \ElsIf{$i \geq d_e$} \Comment{After drift completes}
                \State $I' \gets g(I)$
            \Else \Comment{Gradual drift region}
                \State $p \gets (i - d_s) / w$
                \State $I' \gets g(I)$ with probability $p$, else $I' \gets I$
            \EndIf
            \State $\hat{y} \gets M.\textsc{Predict}(I'.X)$ \Comment{Inference}
            \State $e \gets \textsc{Error}(\hat{y}, I'.y)$ \Comment{Compute error}
            \State $D.\textsc{Update}(e)$ \Comment{Feed detector}
            \If{$D.\textsc{Alarm}()$}
                \State Record detection at position $i$
            \EndIf
            \State $M.\textsc{Train}(I')$ \Comment{Update model}
        \EndFor
        \State Store results for trial $j$
    \EndFor
    \end{algorithmic}
\end{algorithm}

\subsection{Drift Types}\label{app:drift-types}

Algorithms~\ref{alg:drift_y_prior},~\ref{alg:drift_y_swap},~\ref{alg:drift_x_permute}, and~\ref{alg:drift_x_exceed} describe the specific drift types applied in our work using the proposed simulation framework.

\begin{algorithm}[htb]
    \caption{Class Prior Drift Simulation}
    \label{alg:drift_y_prior}
    \begin{algorithmic}[1]
    \Require Instance $I = (x, y)$, Selected label $y_{sel}$, Skip probability $p_{skip}$
    \Ensure Instance $I$ or Null (if skipped)
    \If{$I.y = y_{sel}$}
        \State Generate random value $r \sim \text{Uniform}(0, 1)$
        \If{$r < p_{skip}$}
            \State \Return Null \Comment{Instance is dropped}
        \EndIf
    \EndIf
    \State \Return $I$ \Comment{Instance is kept unchanged}
    \end{algorithmic}
\end{algorithm}

\begin{algorithm}[htb]
    \caption{Class Label Swap Drift Simulation}
    \label{alg:drift_y_swap}
    \begin{algorithmic}[1]
    \Require Instance $I = (x, y)$, Selected label $y_{sel}$, Swap label $y_{swap}$
    \Ensure Transformed instance $I'$
    \If{$I.y = y_{sel}$}
        \State $I' = (I.x, y_{swap})$ \Comment{Create instance with swapped label}
        \State \Return $I'$
    \EndIf
    \State \Return $I$ \Comment{Instance remains unchanged}
    \end{algorithmic}
\end{algorithm}

\begin{algorithm}[htb]
    \caption{Feature Permutation Drift Simulation}
    \label{alg:drift_x_permute}
    \begin{algorithmic}[1]
    \Require Instance $I = (x, y)$, Permutation index vector $\sigma$
    \Ensure Transformed instance $I'$
    \State $x' \leftarrow x[\sigma]$ \Comment{Reorder features according to $\sigma$}
    \State $I' = (x', I.y)$ \Comment{Create instance with permuted features}
    \State \Return $I'$
    \end{algorithmic}
\end{algorithm}

\begin{algorithm}[htb]
    \caption{Feature Distribution Drift Simulation}
    \label{alg:drift_x_exceed}
    \begin{algorithmic}[1]
    \Require Instance $I = (x, y)$, Feature index $j$, Threshold $\tau$
    \Ensure Instance $I$ or Null (if filtered)
    \If{$I.x[j] > \tau$}
        \State \Return Null \Comment{Instance is dropped}
    \EndIf
    \State \Return $I$ \Comment{Instance is kept unchanged}
    \end{algorithmic}
\end{algorithm}

\section{Experiments}

\subsection{Experimental Setup}

Tables~\ref{tab:drift_parameters} and~\ref{tab:hpo_config} summarise the key parameters and configuration pool considered for the experiments.

\begin{table}
\caption{Dataset summary and drift parameters}
\label{tab:drift_parameters}
\resizebox{.45\textwidth}{!}{%
\begin{tabular}{lrrrrr}
\toprule
 & \# Samples & \# Features & \# Classes & Max Delay & Drift Width \\
\midrule
Asfault & 8563 & 62 & 5 & 500 & 500 \\
Covtype & 100000 & 54 & 7 & 2500 & 2500 \\
Electricity & 45312 & 8 & 2 & 2500 & 2500 \\
GasSensor & 13910 & 128 & 6 & 1500 & 1500 \\
NOAA & 18159 & 8 & 2 & 1500 & 1500 \\
Posture & 100000 & 3 & 8 & 2500 & 2500 \\
Rialto & 82250 & 27 & 10 & 2500 & 2500 \\
\bottomrule
\end{tabular}%
}
\end{table}

\begin{table}
\caption{Detector Parameter Search Space}
\label{tab:hpo_config}
\resizebox{\columnwidth}{!}{%
\begin{tabular}{ll>{\raggedright\arraybackslash}p{6cm}}
\toprule
Detector & Parameter & Values \\
\midrule
\multirow[t]{5}{*}{ABCD} & delta\_drift & 0.001, 0.002, 0.0001, 0.01, 0.02 \\
 & delta\_warn & 0.1, 0.01 \\
 & encoding\_factor & 0.3, 0.5, 0.7 \\
 & model\_id & kpca, ae \\
 & num\_splits & 20, 50, 100 \\
\cmidrule(lr){1-3}
\multirow[t]{5}{*}{ABCD(X)} & delta\_drift & 0.001, 0.002, 0.0001, 0.01, 0.02 \\
 & delta\_warn & 0.1, 0.01 \\
 & encoding\_factor & 0.3, 0.5, 0.7 \\
 & model\_id & kpca, ae \\
 & num\_splits & 20, 50, 100 \\
\cmidrule(lr){1-3}
ADWIN & delta & 0.001, 0.002, 0.005, 0.01, 0.0005, 0.0001 \\
\cmidrule(lr){1-3}
\multirow[t]{3}{*}{CUSUM} & min\_n\_instances & 30, 50, 100, 300, 500, 1000, 2000 \\
 & delta & 0.001, 0.002, 0.005, 0.0001, 0.01 \\
 & lambda\_ & 50, 100, 150, 300, 500, 1000, 2000 \\
\cmidrule(lr){1-3}
\multirow[t]{2}{*}{DDM} & min\_n\_instances & 30, 50, 100, 300, 500, 1000, 2000 \\
 & out\_control\_level & 1.75, 2, 2.5, 3.0, 2.25, 2.75, 3.5, 4 \\
\cmidrule(lr){1-3}
\multirow[t]{2}{*}{EWMA} & min\_n\_instances & 50, 100, 300, 500, 1000, 2000, 3000, 5000, 10000 \\
 & lambda\_ & 0.9, 0.01, 0.001, 0.1, 0.005, 0.002, 0.0001 \\
\cmidrule(lr){1-3}
\multirow[t]{3}{*}{GMA} & min\_n\_instances & 30, 50, 100, 300, 500, 1000, 2000 \\
 & lambda\_ & 0.001, 0.002, 0.01, 0.1, 0.5, 1, 2, 3, 5 \\
 & alpha & 0.99, 0.995, 0.9, 0.8, 0.7, 0.5, 0.1, 0.01 \\
\cmidrule(lr){1-3}
\multirow[t]{2}{*}{HDDMA} & drift\_confidence & 0.001, 0.002, 0.005, 0.01, 0.0001 \\
 & test\_type & Two-sided, One-sided \\
\cmidrule(lr){1-3}
\multirow[t]{3}{*}{HDDMW} & drift\_confidence & 0.001, 0.002, 0.005, 0.01, 0.0001 \\
 & test\_type & Two-sided, One-sided \\
 & lambda\_ & 0.05, 0.001, 0.1, 0.01, 0.0001 \\
\cmidrule(lr){1-3}
\multirow[t]{4}{*}{PH} & min\_n\_instances & 30, 50, 100, 300, 500, 1000, 2000 \\
 & delta & 0.001, 0.002, 0.005, 0.01, 0.0001, 0.1 \\
 & lambda\_ & 30, 50, 100, 300, 500, 1000, 2000 \\
 & alpha & 0.99, 0.999, 0.995, 0.9, 0.8, 0.5 \\
\cmidrule(lr){1-3}
\multirow[t]{2}{*}{RDDM} & min\_n\_instances & 30, 50, 100, 300, 500, 1000, 2000 \\
 & drift\_level & 1.9, 2, 2.1, 2.25, 2.5, 3, 1.5, 1.75 \\
\cmidrule(lr){1-3}
\multirow[t]{4}{*}{SEED} & delta & 0.0001, 0.001, 0.01, 0.05, 0.1 \\
 & epsilon\_prime & 0.0025, 0.01, 0.005, 0.0075 \\
 & block\_size & 32, 50, 100, 150, 256 \\
 & alpha & 0.2, 0.3, 0.4, 0.5, 0.6, 0.7, 0.8 \\
\cmidrule(lr){1-3}
\multirow[t]{2}{*}{STEPD} & window\_size & 30, 50, 100, 300, 500, 1000 \\
 & alpha\_drift & 0.001, 0.002, 0.003, 0.005, 0.01, 0.1 \\
\cmidrule(lr){1-3}
STUDD & min\_n\_instances & 500, 1000, 2000, 5000 \\
\cmidrule(lr){1-3}
\bottomrule
\end{tabular}%
}
\end{table}

\subsection{Results}\label{app:results}

This section presents the complete experimental results obtained from evaluating the drift detection methods across all datasets and drift scenarios. Similarly to the main content, in all tables, \textbf{bold} values indicate the best performing method for each drift type, while \underline{underlined} values denote the second-best performance.
Tables~\ref{tab:meanf1} and~\ref{tab:meanf1grd} report the average F1 detection scores for abrupt and gradual drift scenarios, respectively. Tables~\ref{tab:app_precision_abr} and~\ref{tab:app_precision_grd} present the average precision scores for abrupt and gradual drift scenarios, respectively. Tables~\ref{tab:app_recall_abr} and~\ref{tab:app_recall_grd} report the average recall scores for abrupt and gradual drift scenarios, respectively. Table~\ref{tab:synth_metrics} shows the F1 score on synthetic data streams, and Table~\ref{tab:clfs} reports the F1 score across different classifiers. Below, we provide a detailed analysis of each set of results.

Decomposing the F1 scores into precision and recall (Tables~\ref{tab:app_precision_abr}--\ref{tab:app_recall_grd}) reveals different profiles across detectors and clarifies how some methods achieve similar F1 scores through different behaviors.
For example, \texttt{GMA} exhibits a recall-dominated profile: it achieves near-perfect recall (0.97--1.0 across all drift types and abruptness conditions), meaning it detects almost every drift. However, its precision is among the lowest (0.05--0.08), indicating that the majority of its alarms are false positives. This pattern suggests that this method is overly sensitive, triggering alarms frequently regardless of whether a genuine drift has occurred. In practice, however, this kind of behavior leads to excessive and unnecessary model retraining, undermining trust in the detection system. Other detectors such as \texttt{RDDM} exhibit a similar profile, achieving high recall but low precision. 
\texttt{SEED}, \texttt{ABCD}, and \texttt{STEPD} show a relatively balanced precision and recall trade-off across all scenarios.
Finally, under gradual drift conditions, overall scores are lower than for abrupt drifts, but the relative performance of detectors is largely preserved.

\begin{figure}[htb]
    \centering
    \includegraphics[width=.45\textwidth, trim=0cm 0cm 0cm 0cm, clip=TRUE]{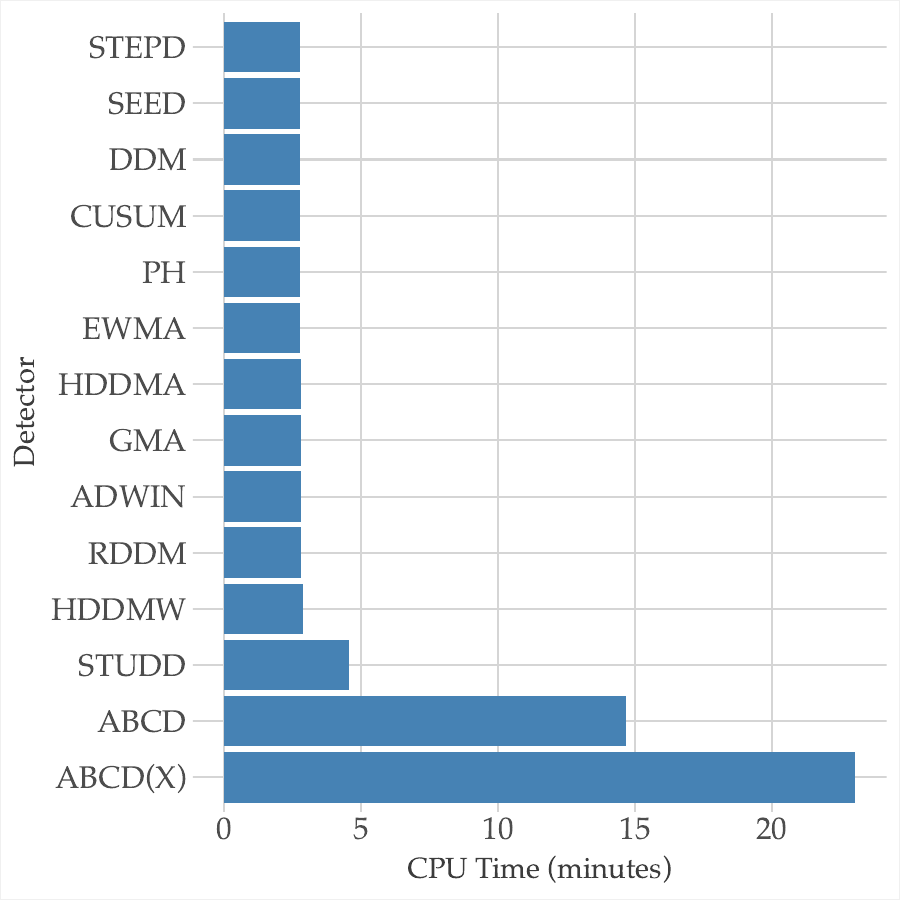}
    \caption{Average execution time of each drift detection method across all datasets}
    \label{fig:plot2_cpu}
    \Description{CPU time.}
\end{figure}

Figure~\ref{fig:plot2_cpu} displays the average execution time (in minutes) for each drift detection method, aggregated across all datasets. The execution time encompasses the complete detection pipeline. This metric complements the main results by providing insight into the computational overhead associated with each detector.
The computational cost varies across methods. Most methods show a similar cost, except for \texttt{ABCD}, \texttt{ABCD(X)}, and \texttt{STUDD} which are more expensive.

Table~\ref{tab:synth_metrics} reports F1 scores on three synthetic data stream generators (Agrawal, SEA, STAGGER), each composed of either abrupt or gradual drifts, using only the error-tracking detectors.
The overall ranking of detectors on synthetic streams broadly corroborates the findings on real-world data: \texttt{SEED}, \texttt{ABCD}, and \texttt{STEPD} achieve the highest F1 scores across most synthetic scenarios. 

Table~\ref{tab:clfs} reports F1 scores for five classifiers---Adaptive Random Forest (ARF), Hoeffding Tree, Naive Bayes, Online Bagging, and OzaBoost---on the Electricity dataset using the label swap drift type. This experiment assesses whether the relative performance of detectors is sensitive to the underlying learning algorithm.
The relative ranking of detectors is largely preserved across classifiers. This stability suggests that the findings from our main experiments, which use the Hoeffding Tree, are not a result of a particular learner's error dynamics.
Absolute F1 scores, however, vary noticeably across classifiers. OzaBoost consistently yields the highest detection scores (e.g., 0.82 for \texttt{SEED}, 0.78 for \texttt{STEPD} and \texttt{HDDMA}), while others tend to produce lower scores.

\begin{table}[htbp]
\caption{Average F1 detection score of drift detectors across different datasets for abrupt drifts}
\label{tab:meanf1}
\resizebox{.45\textwidth}{!}{%
\begin{tabular}{l>{\raggedleft\arraybackslash}p{1.3cm}>{\raggedleft\arraybackslash}p{1.3cm}>{\raggedleft\arraybackslash}p{1.3cm}>{\raggedleft\arraybackslash}p{1.3cm}}
\toprule
Detector & Feature Filtering & Feat. Permutation & Class Prior & Class Swap \\
\midrule
\texttt{ABCD} & 0.11 & 0.51 & 0.16 & 0.25 \\
\texttt{ABCD(X)} & 0.02 & \textbf{0.98} & 0.0 & 0.0 \\
\texttt{ADWIN} & \underline{0.24} & 0.27 & \textbf{0.31} & 0.16 \\
\texttt{CUSUM} & 0.19 & 0.01 & 0.22 & 0.1 \\
\texttt{DDM} & 0.16 & 0.02 & 0.19 & 0.09 \\
\texttt{EWMA} & 0.09 & 0.0 & 0.06 & 0.03 \\
\texttt{GMA} & 0.1 & 0.13 & 0.12 & 0.15 \\
\texttt{HDDMA} & 0.17 & 0.36 & 0.2 & 0.24 \\
\texttt{HDDMW} & 0.18 & 0.25 & 0.19 & 0.21 \\
\texttt{PH} & 0.02 & 0.0 & 0.06 & 0.09 \\
\texttt{RDDM} & 0.2 & 0.04 & 0.15 & 0.1 \\
\texttt{SEED} & \underline{0.24} & \underline{0.61} & \underline{0.3} & \textbf{0.37} \\
\texttt{STEPD} & \textbf{0.26} & 0.53 & 0.28 & \underline{0.33} \\
\texttt{STUDD} & 0.17 & 0.27 & 0.1 & 0.11 \\
\bottomrule
\end{tabular}%
}
\end{table}

\begin{table}[htbp]
\caption{Average F1 detection score of drift detectors across different datasets for gradual drifts}
\label{tab:meanf1grd}
\resizebox{.45\textwidth}{!}{%
\begin{tabular}{l>{\raggedleft\arraybackslash}p{1.3cm}>{\raggedleft\arraybackslash}p{1.3cm}>{\raggedleft\arraybackslash}p{1.3cm}>{\raggedleft\arraybackslash}p{1.3cm}}
\toprule
Detector & Feature Filtering & Feat. Permutation & Class Prior & Class Swap \\
\midrule
\texttt{ABCD} & 0.03 & 0.33 & 0.07 & 0.18 \\
\texttt{ABCD(X)} & 0.0 & \textbf{0.7} & 0.0 & 0.0 \\
\texttt{ADWIN} & 0.1 & 0.04 & 0.15 & 0.08 \\
\texttt{CUSUM} & 0.11 & 0.01 & 0.14 & 0.08 \\
\texttt{DDM} & \underline{0.13} & 0.06 & 0.15 & 0.1 \\
\texttt{EWMA} & 0.04 & 0.0 & 0.04 & 0.02 \\
\texttt{GMA} & \underline{0.13} & 0.13 & 0.14 & 0.15 \\
\texttt{HDDMA} & 0.07 & 0.1 & 0.13 & 0.06 \\
\texttt{HDDMW} & 0.11 & 0.14 & 0.13 & 0.16 \\
\texttt{PH} & 0.05 & 0.01 & 0.1 & 0.05 \\
\texttt{RDDM} & \underline{0.13} & 0.13 & 0.14 & 0.14 \\
\texttt{SEED} & 0.11 & \underline{0.46} & \underline{0.17} & \texttt{0.29} \\
\texttt{STEPD} & \underline{0.13} & 0.45 & \textbf{0.2} & \underline{0.28} \\
\texttt{STUDD} & \textbf{0.14} & 0.15 & 0.13 & 0.1 \\
\bottomrule
\end{tabular}%
}
\end{table}

\begin{table}[htbp]
\caption{Average precision score of drift detectors across different datasets for abrupt drifts}
\label{tab:app_precision_abr}
\resizebox{.45\textwidth}{!}{%
\begin{tabular}{l>{\raggedleft\arraybackslash}p{1.3cm}>{\raggedleft\arraybackslash}p{1.3cm}>{\raggedleft\arraybackslash}p{1.3cm}>{\raggedleft\arraybackslash}p{1.3cm}}
\toprule
Detector & Feature Filtering & Feat. Permutation & Class Prior & Class Swap \\
\midrule
\texttt{ABCD} & 0.11 & 0.43 & 0.17 & 0.23 \\
\texttt{ABCD(X)} & 0.03 & \textbf{1.0} & 0.0 & 0.0 \\
\texttt{ADWIN} & 0.18 & 0.19 & \underline{0.26} & 0.1 \\
\texttt{CUSUM} & 0.18 & 0.01 & 0.21 & 0.08 \\
\texttt{DDM} & 0.14 & 0.02 & 0.16 & 0.06 \\
\texttt{EWMA} & 0.09 & 0.0 & 0.08 & 0.07 \\
\texttt{GMA} & 0.05 & 0.07 & 0.06 & 0.08 \\
\texttt{HDDMA} & 0.19 & 0.27 & 0.19 & 0.17 \\
\texttt{HDDMW} & 0.16 & 0.17 & 0.15 & 0.15 \\
\texttt{PH} & 0.04 & 0.0 & 0.19 & 0.16 \\
\texttt{RDDM} & 0.16 & 0.03 & 0.11 & 0.07 \\
\texttt{SEED} & \textbf{0.24} & \underline{0.49} & \texttt{0.27} & \textbf{0.32} \\
\texttt{STEPD} & \underline{0.23} & 0.4 & 0.24 & \underline{0.26} \\
\texttt{STUDD} & 0.11 & 0.19 & 0.06 & 0.07 \\
\bottomrule
\end{tabular}%
}
\end{table}

\begin{table}[htbp]
\caption{Average precision score of drift detectors across different datasets for gradual drifts}
\label{tab:app_precision_grd}
\resizebox{.45\textwidth}{!}{%
\begin{tabular}{l>{\raggedleft\arraybackslash}p{1.3cm}>{\raggedleft\arraybackslash}p{1.3cm}>{\raggedleft\arraybackslash}p{1.3cm}>{\raggedleft\arraybackslash}p{1.3cm}}
\toprule
Detector & Feature Filtering & Feat. Permutation & Class Prior & Class Swap \\
\midrule
\texttt{ABCD} & 0.03 & 0.26 & 0.08 & \underline{0.17} \\
\texttt{ABCD(X)} & 0.0 & \textbf{0.66} & 0.0 & 0.0 \\
\texttt{ADWIN} & 0.07 & 0.03 & 0.1 & 0.05 \\
\texttt{CUSUM} & \underline{0.1} & 0.02 & \underline{0.13} & 0.08 \\
\texttt{DDM} & 0.07 & 0.04 & 0.08 & 0.06 \\
\texttt{EWMA} & 0.05 & 0.0 & 0.05 & 0.04 \\
\texttt{GMA} & 0.07 & 0.07 & 0.07 & 0.08 \\
\texttt{HDDMA} & 0.06 & 0.09 & \underline{0.13} & 0.05 \\
\texttt{HDDMW} & 0.07 & 0.08 & 0.07 & 0.09 \\
\texttt{PH} & 0.06 & 0.01 & 0.12 & 0.05 \\
\texttt{RDDM} & 0.07 & 0.07 & 0.07 & 0.08 \\
\texttt{SEED} & 0.09 & \underline{0.33} & \underline{0.13} & \texttt{0.23} \\
\texttt{STEPD} & \textbf{0.11} & \underline{0.33} & \texttt{0.16} & \textbf{0.23} \\
\texttt{STUDD} & 0.09 & 0.1 & 0.09 & 0.06 \\
\bottomrule
\end{tabular}%
}
\end{table}

\begin{table}[htbp]
\caption{Average recall score of drift detectors across different datasets for abrupt drifts}
\label{tab:app_recall_abr}
\resizebox{.45\textwidth}{!}{%
\begin{tabular}{l>{\raggedleft\arraybackslash}p{1.3cm}>{\raggedleft\arraybackslash}p{1.3cm}>{\raggedleft\arraybackslash}p{1.3cm}>{\raggedleft\arraybackslash}p{1.3cm}}
\toprule
Detector & Feature Filtering & Feat. Permutation & Class Prior & Class Swap \\
\midrule
\texttt{ABCD} & 0.14 & 0.7 & 0.17 & 0.37 \\
\texttt{ABCD(X)} & 0.01 & \underline{0.96} & 0.0 & 0.0 \\
\texttt{ADWIN} & 0.4 & 0.55 & \underline{0.51} & 0.47 \\
\texttt{CUSUM} & 0.23 & 0.02 & 0.26 & 0.18 \\
\texttt{DDM} & 0.23 & 0.04 & 0.29 & 0.21 \\
\texttt{EWMA} & 0.1 & 0.0 & 0.08 & 0.05 \\
\texttt{GMA} & \textbf{0.97} & \textbf{1.0} & \textbf{1.0} & \texttt{1.0} \\
\texttt{HDDMA} & 0.23 & 0.66 & 0.3 & 0.45 \\
\texttt{HDDMW} & 0.33 & 0.67 & 0.45 & 0.54 \\
\texttt{PH} & 0.02 & 0.0 & 0.04 & 0.07 \\
\texttt{RDDM} & 0.33 & 0.11 & 0.32 & 0.29 \\
\texttt{SEED} & 0.29 & 0.93 & 0.36 & \underline{0.58} \\
\texttt{STEPD} & 0.36 & 0.92 & 0.39 & \underline{0.58} \\
\texttt{STUDD} & \underline{0.48} & 0.75 & 0.38 & 0.41 \\
\bottomrule
\end{tabular}%
}
\end{table}

\begin{table}[htbp]
\caption{Average recall score of drift detectors across different datasets for gradual drifts}
\label{tab:app_recall_grd}
\resizebox{.45\textwidth}{!}{%
\begin{tabular}{l>{\raggedleft\arraybackslash}p{1.3cm}>{\raggedleft\arraybackslash}p{1.3cm}>{\raggedleft\arraybackslash}p{1.3cm}>{\raggedleft\arraybackslash}p{1.3cm}}
\toprule
Detector & Feature Filtering & Feat. Permutation & Class Prior & Class Swap \\
\midrule
\texttt{ABCD} & 0.04 & 0.5 & 0.08 & 0.28 \\
\texttt{ABCD(X)} & 0.0 & 0.76 & 0.0 & 0.0 \\
\texttt{ADWIN} & 0.24 & 0.1 & 0.34 & 0.27 \\
\texttt{CUSUM} & 0.14 & 0.02 & 0.17 & 0.11 \\
\texttt{DDM} & \underline{0.89} & 0.69 & \underline{0.89} & 0.78 \\
\texttt{EWMA} & 0.06 & 0.01 & 0.06 & 0.03 \\
\texttt{GMA} & \textbf{1.0} & \textbf{1.0} & \textbf{1.0} & \textbf{1.0} \\
\texttt{HDDMA} & 0.09 & 0.17 & 0.15 & 0.11 \\
\texttt{HDDMW} & 0.65 & 0.78 & 0.66 & \underline{0.81} \\
\texttt{PH} & 0.05 & 0.01 & 0.09 & 0.05 \\
\texttt{RDDM} & \textbf{1.0} & \underline{0.99} & \texttt{1.0} & \textbf{1.0} \\
\texttt{SEED} & 0.19 & 0.87 & 0.27 & 0.59 \\
\texttt{STEPD} & 0.19 & 0.84 & 0.28 & 0.54 \\
\texttt{STUDD} & 0.43 & 0.61 & 0.37 & 0.41 \\
\bottomrule
\end{tabular}%
}
\end{table}

\begin{table}[htbp]
\caption{F1 scores on synthetic data streams (Agrawal, SEA, STAGGER) composed of abrupt or gradual drifts.}
\label{tab:synth_metrics}
\resizebox{.45\textwidth}{!}{%
\begin{tabular}{lrrr|rrr}
\toprule
Mode & \multicolumn{3}{c}{ABRUPT} & \multicolumn{3}{c}{GRADUAL} \\
Stream & Agrawal & SEA & STAGGER & Agrawal & SEA & STAGGER \\
\midrule
\texttt{ABCD} & \underline{0.57} & 0.07 & \textbf{0.40} & \underline{0.60} & 0.33 & 0.44 \\
\texttt{ADWIN} & 0.26 & 0.04 & 0.04 & 0.54 & 0.54 & 0.08 \\
\texttt{CUSUM} & 0.12 & 0.00 & 0.04 & 0.17 & 0.22 & 0.03 \\
\texttt{DDM} & 0.10 & 0.00 & 0.04 & 0.34 & 0.36 & 0.11 \\
\texttt{EWMA} & 0.00 & 0.00 & 0.00 & 0.00 & 0.00 & 0.00 \\
\texttt{GMA} & 0.18 & 0.19 & 0.16 & 0.40 & 0.39 & 0.45 \\
\texttt{HDDMA} & 0.22 & 0.05 & 0.12 & 0.32 & 0.26 & 0.16 \\
\texttt{HDDMW} & 0.16 & 0.09 & 0.05 & 0.45 & 0.39 & \textbf{0.50} \\
\texttt{PH} & 0.06 & 0.00 & 0.00 & 0.14 & 0.07 & 0.03 \\
\texttt{RDDM} & 0.15 & 0.08 & 0.04 & 0.40 & 0.43 & 0.41 \\
\texttt{SEED} & \textbf{0.58} & \underline{0.46} & \underline{0.35} & \textbf{0.61} & \textbf{0.63} & \underline{0.47} \\
\texttt{STEPD} & 0.49 & \textbf{0.49} & 0.29 & \textbf{0.61} & \underline{0.61} & 0.42 \\
\bottomrule
\end{tabular}%
}
\end{table}

\begin{table}
\caption{F1 scores of each detector for different classifiers using the Electricity data stream and the label swapping drift type.}
\label{tab:clfs}
\resizebox{.5\textwidth}{!}{%
\begin{tabular}{llllll}
\toprule
Learner & ARF & HoeffdingTree & NaiveBayes & OnlineBagging & OzaBoost \\
\midrule
\texttt{ABCD} & 0.51 & 0.33 & \textbf{0.53} & 0.35 & 0.52 \\
\texttt{ADWIN} & 0.23 & 0.19 & 0.44 & 0.22 & 0.68 \\
\texttt{CUSUM} & 0.0 & 0.1 & 0.37 & 0.01 & 0.6 \\
\texttt{DDM} & 0.03 & 0.06 & 0.25 & 0.06 & 0.4 \\
\texttt{EWMA} & 0.0 & 0.0 & 0.0 & 0.0 & 0.0 \\
\texttt{GMA} & 0.2 & 0.18 & 0.15 & 0.2 & 0.17 \\
\texttt{HDDMA} & 0.56 & \underline{0.51} & 0.41 & 0.47 & \underline{0.78} \\
\texttt{HDDMW} & \textbf{0.62} & 0.42 & 0.22 & 0.46 & 0.63 \\
\texttt{PH} & 0.0 & 0.02 & 0.06 & 0.0 & \underline{0.78} \\
\texttt{RDDM} & 0.06 & 0.12 & 0.18 & 0.08 & 0.3 \\
\texttt{SEED} & 0.51 & \textbf{0.54} & \underline{0.5} & \underline{0.53} & \textbf{0.82} \\
\texttt{STEPD} & \underline{0.57} & \textbf{0.54} & 0.46 & \textbf{0.56} & \underline{0.78} \\
\bottomrule
\end{tabular}%
}
\end{table}

\end{document}